\RequirePackage{fix-cm}

\documentclass[twocolumn]{svjour3}          % twocolumn
\smartqed  % flush right qed marks, e.g. at end of proof
\usepackage{graphicx}

\journalname{IJCV}

\usepackage{amsmath}
\usepackage{amssymb}
\usepackage{mathtools}
\usepackage{listings}
\usepackage{booktabs} %
\usepackage{multirow}
\usepackage{amsfonts}
\usepackage{nicefrac}
\usepackage{microtype}
\usepackage{xspace}
\usepackage{framed}
\usepackage{color, colortbl}
\usepackage{subfigure}
\usepackage{caption}
\usepackage{wrapfig}
\usepackage{xcolor} 
\usepackage[colorlinks]{hyperref}
\usepackage[misc]{ifsym}
\usepackage{multicol}

\newcommand{\cmark}{$\surd$}
\newcommand{\xmarkg}{$\times$}

\def \ours {CAE*\xspace} 
\def \oursdvae {CAE\xspace} 
\definecolor{cornflowerblue}{RGB}{100, 149, 237}

\newcommand*{\rowstyle}[1]{
  \gdef\@rowstyle{#1}%
  \@rowstyle\ignorespaces%
}
\newcolumntype{=}{
  >{\gdef\@rowstyle{}}%
}
\newcolumntype{+}{
  >{\@rowstyle}%
}
\makeatother

\begin{document}

\title{Context Autoencoder for Self-Supervised Representation Learning}

\author{Xiaokang Chen$^1$ \and Mingyu Ding$^{2,3}$ \and Xiaodi Wang$^4$ \and Ying Xin$^4$ \and Shentong Mo$^4$ \and Yunhao Wang$^4$ \and Shumin Han$^4$ \and Ping Luo$^2$ \and \\ Gang Zeng$^1$ \and Jingdong Wang$^4$}

\authorrunning{Xiaokang Chen et al.}

\institute{$^1$Peking University \\  $^2$University of Hong Kong \\ $^3$UC Berkeley \\ $^4$Baidu \\ \Letter { wangjingdong@outlook.com}}

\date{Received: date / Accepted: date}

\maketitle

\begin{abstract}
We present a novel masked image modeling
(MIM) approach,
context autoencoder (CAE),
for self-supervised representation pretraining.
We pretrain an encoder
by making predictions
in the encoded representation space.
The pretraining tasks include
two tasks: masked representation prediction - predict the representations
for the masked patches,
and masked patch reconstruction - reconstruct the masked patches.
The network is an encoder-regressor-decoder architecture:
the encoder takes the visible patches as input;
the regressor predicts the representations of the masked patches,
which are expected to be aligned
with the representations
computed from the encoder,
using the representations of visible patches and the positions of visible and masked patches;
the decoder reconstructs the masked patches
from the predicted encoded representations.
The CAE design encourages 
the separation of learning the encoder (representation) 
from completing the pertaining tasks: masked representation prediction and masked patch reconstruction tasks,
and making predictions in the encoded representation space
empirically shows
the benefit
to representation learning.
We demonstrate the effectiveness of our CAE
through 
superior transfer 
performance in downstream tasks:
semantic segmentation, object detection and instance segmentation, and classification.
The code will be available at~\url{https://github.com/Atten4Vis/CAE}.
\end{abstract}

\keywords{Self-Supervised Representation Learning, Masked Image Modeling, Context Autoencoder}

\section{Introduction}\label{sec1}

We study the masked image modeling (MIM) task 
for self-supervised representation learning.
It aims to learn an encoder
through masking some patches of the
input image and 
making predictions for the masked patches
from the visible patches.
It is expected that
the resulting encoder 
pretrained through solving the MIM task 
is able to extract the patch representations
taking on semantics
that are transferred to solving downstream tasks.

\begin{figure}[t]
\centering
\centerline{\includegraphics[width=0.98\columnwidth]{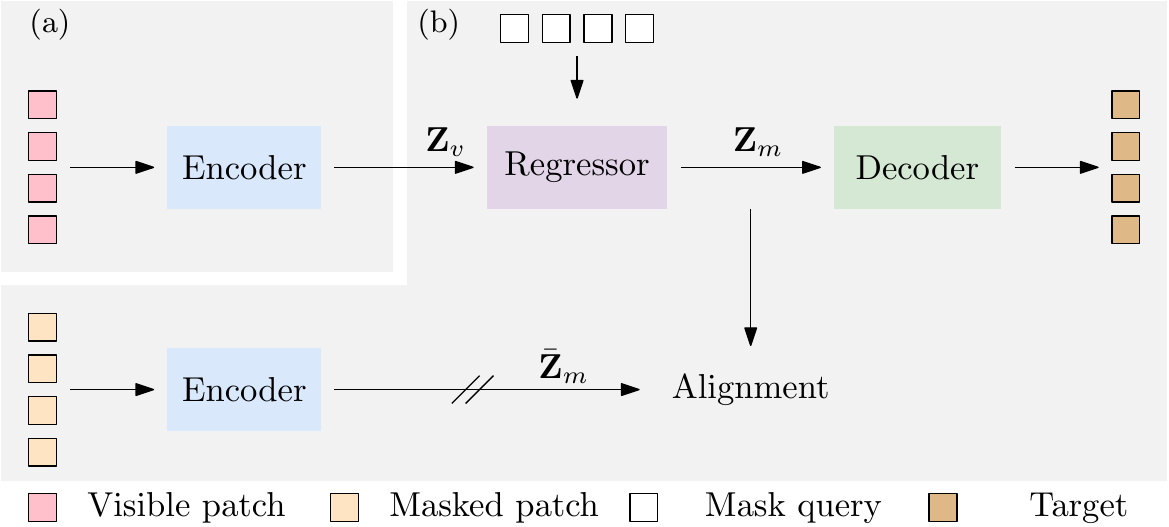}}
\caption{The pipeline of context autoencoder.
Our approach 
(a) feeds visible patches into
the encoder
and extracts their representations $\mathbf{Z}_v$
and then (b) completes the pretext tasks:
predict
the representations $\mathbf{Z}_m$
of the masked patches
from the visible patches
in the encoded representation space
through 
latent contextual regressor and prediction alignment,
and reconstruct the masked patches from the predicted 
representations $\mathbf{Z}_m$
of masked patches.
The pretrained encoder in (a) 
is applied to downstream tasks
by simply replacing the pretext task part
(b)
with the downstream task part.
$//$ means stop gradient.
}
\label{fig:CAE}
\end{figure}

The typical MIM methods,
such as 
BEiT~\cite{bao2021beit},
the method studied in the ViT paper~\cite{DosovitskiyB0WZ21},
and iBoT~\cite{zhou2021ibot},
use a single ViT architecture 
to solve the pretraining task i.e., reconstructing the patch tokens or the pixel colors.
These methods mix the two tasks: learning the encoder (representation)
and reconstructing the masked patch.
The subsequent method,
masked autoencoder (MAE)~\cite{he2021masked}
adopts an encoder-decoder architecture,
partially decoupling the two tasks.
As a result,
the representation quality is limited.
Most previous methods, except iBoT~\cite{zhou2021ibot}, lack an explicit modeling
between encoded representations 
of visible patches and masked patches.

We present a context autoencoder (CAE) approach,
illustrated in Figure~\ref{fig:CAE},
for improving the encoding quality.
We pretrain the encoder
through making predictions for the masked patches
in the encoded representation space.
The pretraining task
is a combination of masked representation prediction
and masked patch reconstruction.
the pretraining network
is an encoder-regressor-decoder architecture.
The encoder
takes only the visible patches as input
and learns the representations
only for the visible patches.
The regressor
predicts the masked patch representations,
which is expected to
be aligned with the representations
of the masked patches computed from the encoder,
from the visible patch representations.
The decoder reconstructs 
the masked patches
from the predicted masked patch representations
without receiving the representations
of the visible patches.

The prediction in the encoded representation space
from the visible patches
to the masked patches
generates a plausible semantic guess
for the masked patches,
which lies in 
the same semantic space
for the visible patches.
We assume that 
the prediction is easier
if the encoded representations take higher semantics
and 
that the accurate prediction encourages
that the encoded representations
take on a larger extent of semantics,
empirically validated
by the experiments.

The CAE design also encourages 
the separation
of learning the encoder
and completing the pretraining tasks:
the responsibility
of representation learning is 
mainly
taken by the encoder
and the encoder is only for representation learning.
The reasons include: the encoder in the top stream in Figure~\ref{fig:CAE}
operates only on visible patches,
only focusing on learning semantic representations;
the regression is done
on the encoded representation space,
as a mapping between the representations
of the visible patches 
and the masked patches;
the decoder operates
only on the predicted representations
of the masked patches.

We present the empirical performance of our approach
on downstream tasks, semantic segmentation, object detection and instance segmentation, and classification.
The results show that our approach outperforms supervised pretraining, contrastive self-supervised pretraining, and other MIM methods.

\section{Related Work}
Self-supervised representation learning
has been widely studied 
in computer vision
, including:
context prediction~\cite{CarlDoersch2015UnsupervisedVR,tian2021semantic},
clustering-based methods~\cite{xie2016unsupervised,yang2016joint,caron2018deep,asano2019self,zhuang2019local,huang2019unsupervised,caron2019unsupervised,PriyaGoyal2021SelfsupervisedPO},
contrastive self-supervised learning~\cite{li2020prototypical,AaronvandenOord2018RepresentationLW,henaff2020data,wang2022repre},
instance discrimination~\cite{dosovitskiy2014discriminative,dosovitskiy2015discriminative},
image discretization~\cite{gidaris2020learning,gidaris2020online},
masked image modeling~\cite{li2021mst,fang2022corrupted,tian2022beyond},
and information maximization~\cite{ermolov2021whitening,zbontar2021barlow,bardes2021vicreg}.
The following mainly reviews closely-related methods.

\vspace{1mm}
\noindent\textbf{Autoencoding.}
Traditionally, autoencoders were used for dimensionality reduction or feature learning~\cite{phdthesis_LeCun,gallinari1987memoires,hinton1994autoencoders,hinton2006reducing,ranzato2007efficient,vincent2008extracting,kingma2013auto}.
The denoising autoencoder (DAE) is an autoencoder that receives a corrupted data point as input and is trained to estimate the original, uncorrupted data point as its output.
The variants or modifications
of DAE were adopted
for self-supervised representation learning,
e.g., corruption by masking pixels~\cite{VincentLLBM10,pathak2016context,chen2020generative},
removing color channels~\cite{zhang2016colorful},
shuffling image patches~\cite{noroozi2016unsupervised}, denoising pixel-level noise~\cite{atito2021sit}
and so on.

\vspace{1mm}
\noindent\textbf{Contrastive self-supervised learning.}
Contrastive self-supervised learning,
referring 
in this paper
to the self-supervised approaches
comparing random views
with contrastive loss
or simply MSE loss
that are related
as shown in~\cite{GarridoCBNL22},
has been popular
for self-supervised representation learning~\cite{ChenK0H20,He0WXG20,YonglongTian2020WhatMF,ChenXH21,grill2020bootstrap,CaronTMJMBJ21,chen2021exploring,caron2020unsupervised_swav,wu2018unsupervised,XiangyuPeng2022CraftingBC}.
The basic idea is to 
maximize the similarity
between the views augmented
from the same image
and optionally minimize
the similarity
between the views
augmented from different images.
Random cropping
is an important augmentation scheme,
and thus
typical contrastive self-supervised learning methods 
(e.g., MoCo v3) tend to learn knowledge
mainly from the central regions of the original images.
Some dense variants~\cite{wang2021dense,xie2021propagate} 
eliminate the tendency 
in a limited degree
by
considering an extra contrastive loss
with dense patches.

\vspace{1mm}
\noindent\textbf{Masked image modeling.}
Motivated by BERT for masked language modeling~\cite{DevlinCLT19}, 
the method studied in~\cite{DosovitskiyB0WZ21} and BEiT~\cite{bao2021beit} use the ViT structure
to solve the masked image modeling task,
e.g., estimate the pixels
or the discrete tokens.
The follow-up work, iBOT~\cite{zhou2021ibot},
combines the MIM method (BEiT) and 
a contrastive self-supervised approach (DINO~\cite{CaronTMJMBJ21}).
But they do not have explicitly an encoder 
for representation learning 
or a decoder for pretraining task completion,
and the ViT structure is essentially a mixture of encoder and decoder,
limiting the representation learning quality.

Several subsequent MIM methods
are developed to improve the encoder quality, 
such as designing pretraining architectures:
Masked Autoencoder (MAE)~\cite{he2021masked},
SplitMask~\cite{el2021large},
and Simple MIM (SimMIM)~\cite{xie2021simmim};
adopting new reconstruction targets: 
Masked Feature Prediction (MaskFeat)~\cite{wei2021masked},
Perceptual Codebook for BEiT (PeCo)~\cite{dong2021peco},
and
data2vec~\cite{BaevskiHXBGA22}.
The technical report~\footnote{\url{https://arxiv.org/abs/2202.03026}} of our approach
was initially published as an arXiv paper~\cite{CAE2022},
and was concurrent to data2vec~\cite{BaevskiHXBGA22}, MAE~\cite{he2021masked},
and other methods, such as~\cite{el2021large,xie2021simmim}.
After that, MIM methods have developed rapidly,
e.g., extended to frequency/semantic domain~\cite{xie2022masked,liu2022devil,wei2022mvp,li2022mc},  
combined with contrastive self-superivsed learning~\cite{tao2022siamese,jing2022masked,yi2022masked,huang2022contrastive}, 
efficient pretraining~\cite{zhang2022hivit,huang2022green,chen2022efficient}, 
mask strategy design~\cite{kakogeorgiou2022hide,li2022semmae,li2022uniform},
scalability of MIM~\cite{xie2022data},
and interpretation of MIM~\cite{xie2022revealing,li2022architecture,kong2022understanding}.

The core idea of our approach 
is making predictions in the encoded representation space.
We jointly solve two pretraining tasks:
masked representation prediction - predict the representations for the masked patches, 
where the representations lie in the representation space output from the encoder,
and masked patch reconstruction - reconstruct the masked patches.

\begin{figure}[t]
\footnotesize`
\centering
\centerline{\includegraphics[width=0.98\columnwidth]{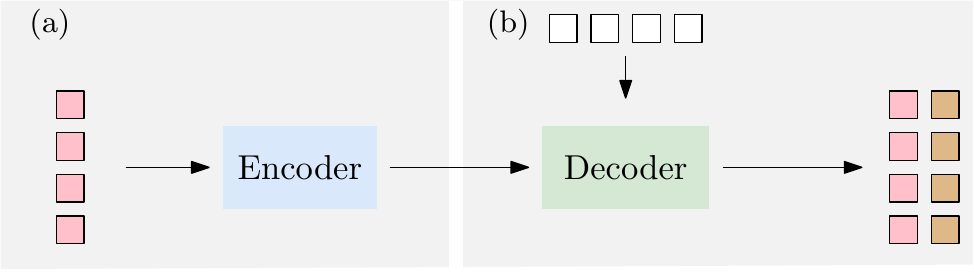}}\\
\centerline{\includegraphics[width=0.98\columnwidth]{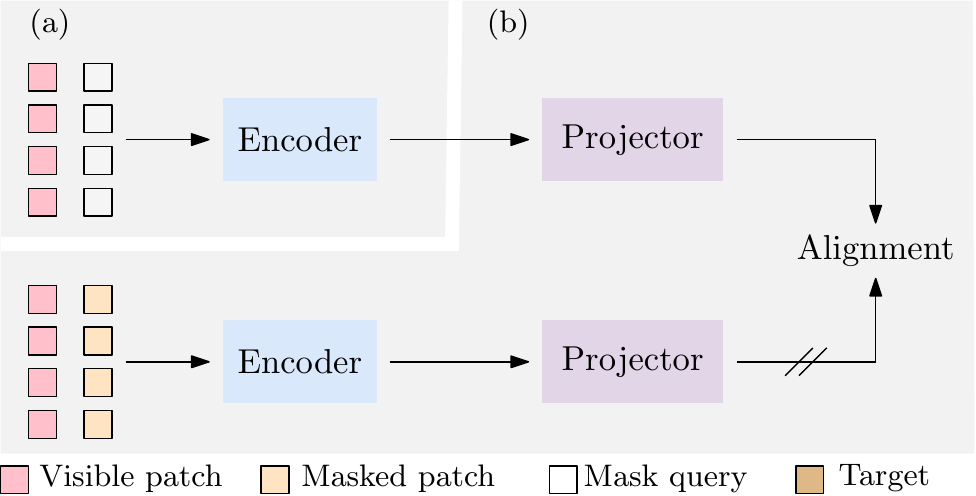}}
\caption{
The pipeline of MAE (top), and the MIM part of iBoT (bottom).
The centering module is not depicted in the bottom stream. The pretrained encoder in (a) 
is applied to downstream tasks
by simply replacing the pretext task part
(b)
with the downstream task part. $//$ means stop gradient.
}
\label{fig:MAEiBoT}
\end{figure}

Our approach is clearly different from MAE~\cite{he2021masked} (Figure~\ref{fig:MAEiBoT} (top)).
Our approach introduces an extra pretraining task, masked representation prediction,
and encourages the separation 
of two roles: learning the encoder and completing pretraining tasks;
in contrast,
MAE partially mixes the two roles,
and has no explicit prediction of masked patch representations.

On the other hand,
our approach differs from data2vec~\cite{BaevskiHXBGA22}
and iBoT~\cite{zhou2021ibot} (Figure~\ref{fig:MAEiBoT} (bottom)).
Similar to BEiT,
in data2vec and iBoT,
there is no explicit module separation  
of learning the encoder and estimating the mask patch representations,
and 
the target representations are formed from the full view (as the teacher)
with 
the same network as
the student network for processing the masked view and predicting the masked patch representations (except a centering process in iBoT for the teacher following DINO).
In contrast, our approach is simple:
form the target representations 
merely from the output of the encoder,
and the encoder-regressor design is straightforward
and explainable:
the regressor predicts the representations of masked patches to match the representations 
computed directly from the encoder.

\section{Approach}
\label{sec:cae}

\subsection{Architecture}
Our context autoencoder (CAE)
is a masked image modeling approach.
The network shown in Figure~\ref{fig:CAE}
is an encoder-regressor-decoder architecture.
The key is to
make predictions from visible patches
to masked patches
in the encoded representation space.
The pretraining tasks include:
masked representation prediction
and masked patch reconstruction. 

We randomly split an image
into two sets of patches:
visible patches $\mathbf{X}_v$
and masked patches $\mathbf{X}_m$.
The encoder takes the visible patches as input;
the regressor predicts the representations of the masked patches,
which are expected to be aligned
with the representations
computed from the encoder,
from the representations of the visible patches conditioned on the positions of masked patches;
the decoder reconstructs the masked patches
from the predicted encoded representations.

\vspace{1mm}
\noindent\textbf{Encoder.}
The encoder $\mathcal{F}$ maps the visible patches
$\mathbf{X}_v$
to the latent representations\
$\mathbf{Z}_v$.
It only handles the visible patches.
We use the ViT to form our encoder.
It first embeds the visible patches 
by linear projection
as patch embeddings,
and adds the positional embeddings $\mathbf{P}_v$.
Then it sends the combined embeddings
into a sequence of transformer blocks
that are based on {self-attention}, 
generating $\mathbf{Z}_v$.

\vspace{1mm}
\noindent\textbf{Regressor.}
The latent contextual regressor $\mathcal{H}$
predicts the latent representations
$\mathbf{Z}_m$
for the masked patches
from the latent representations
$\mathbf{Z}_v$
of the visible patches output from the encoder
conditioned on the positions of
the masked patches.
We form the latent contextual regressor $\mathcal{H}$
using a series of transformer blocks
that are based on {cross-attention}.

The initial queries $\mathbf{Q}_m$,
called mask queries,
are mask tokens
that are learned as model parameters
and are the same for all the masked patches. 
The keys and the values are the same before linear projection and
consist of 
the visible patch representations $\mathbf{Z}_v$
and the output of the previous cross-attention layer
(mask queries for the first cross-attention layer).
The corresponding positional embeddings
of the masked patches
are considered
when computing the cross-attention weights
between the queries and the keys.
In this process,
the latent representations $\mathbf{Z}_v$ of the visible patches 
are not updated.

\vspace{1mm}
\noindent\textbf{Decoder.}
The decoder $\mathcal{G}$ maps the latent representations $\mathbf{Z}_m$
of the masked patches
to some forms of masked patches,
$\mathbf{Y}_m$.
The decoder, similar to the encoder,
is a stack of transformer blocks
that are based on {self-attention},
followed by a linear layer predicting the targets.
The decoder only receives
the latent representations
of the masked patches
(the output of the latent contextual regressor),
and the positional embeddings of the masked patches
as input
without directly using the information of
the visible patches.

\subsection{Objective Function}

\begin{figure}[t]
\centering
\includegraphics[width=0.48\linewidth]{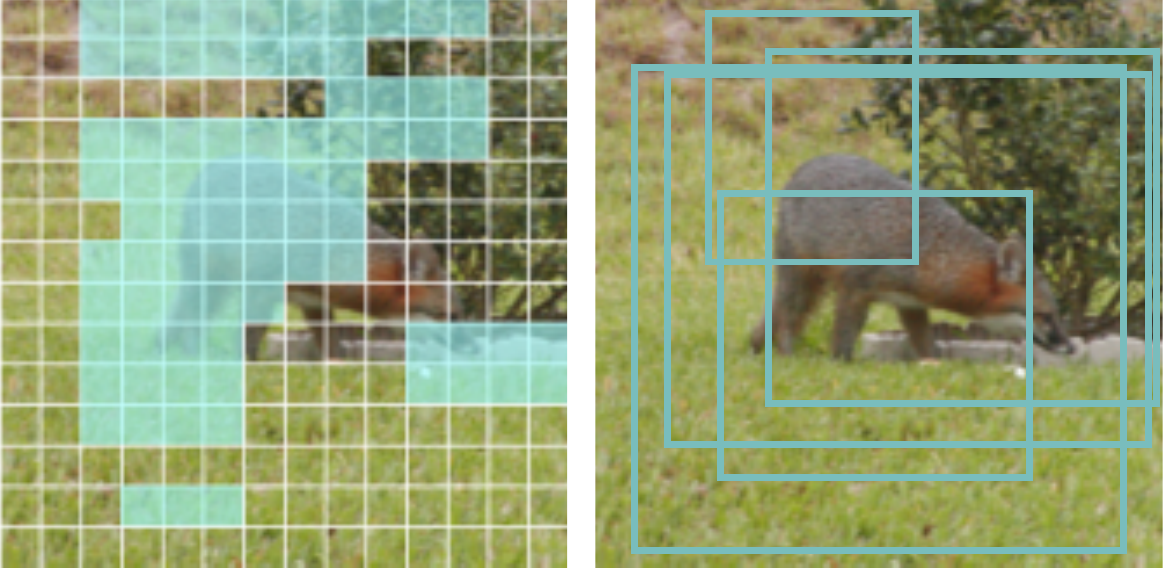}~
\includegraphics[width=0.48\linewidth]{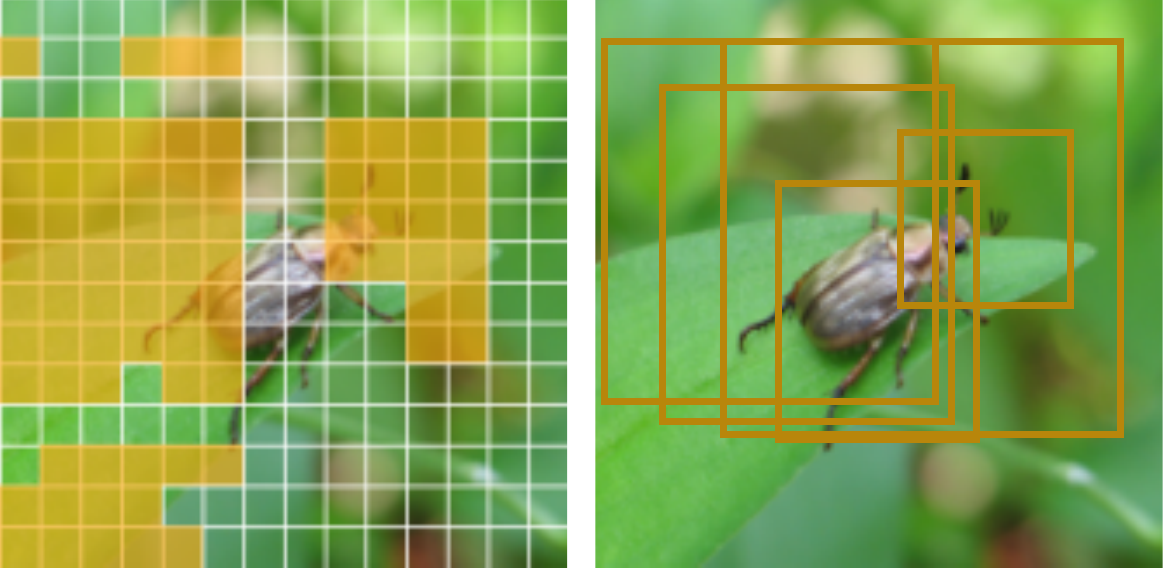}
\caption{ Illustration
of random block-wise sampling ($1$st and $3$rd images)
and random cropping ($2$nd and $4$th images). The colored regions are masked regions. The boxes correspond to cropped regions.
Random block-wise sampling is used in our approach.
Random cropping
is a key data-augmentation scheme
for contrastive self-supervised pretraining. 
}
\label{fig:maskingandcropping}
\end{figure}

\noindent\textbf{Masking.}
Following BEiT~\cite{bao2021beit}, we adopt the random block-wise masking strategy 
(illustrated in Figure~\ref{fig:maskingandcropping}) to split the input image into two sets of patches,
visible and masked patches. For each image, $98$ of $196$ ($14 \times 14$) patches are masked.

\vspace{1mm}
\noindent\textbf{Targets.}
The targets $\bar{\mathbf{Z}}_m$
for the representations
of the masked patches
are formed as follows.
We feed the masked patches $\mathbf{X}_m$
into the encoder,
which is the same as the one for encoding visible patches,
and generate the representations $\bar{\mathbf{Z}}_m$
of the masked patches
as the representation targets.

The targets $\bar{\mathbf{Y}}_m$
for the patch reconstruction 
are formed by 
the discrete tokenizer,
e.g.,
the tokenizer trained with d-VAE
on ImageNet-$1$K without using the labels
or the DALL-E tokenizer
(trained with d-VAE on $400$M images)~\cite{RameshPGGVRCS21}
used in BEiT~\cite{bao2021beit}.
The input image is fed into the tokenizer,
assigning a discrete token to each patch
for forming the reconstruction targets $\bar{\mathbf{Y}}_m$.

\vspace{1mm}
\noindent\textbf{Loss function.}
The loss function 
consists of
a reconstruction loss:
$\ell_y(\mathbf{Y}_m, \bar{\mathbf{Y}}_m)$,
and an alignment loss:
$\ell_z(\mathbf{Z}_m, \bar{\mathbf{Z}}_m)$,
corresponding to masked patch reconstruction and masked representation prediction, respectively.
The whole loss is a weighted sum:
\begin{align}
\quad \quad \quad \quad \quad \ell_y(\mathbf{Y}_m, \bar{\mathbf{Y}}_m)
+ 
\lambda~\ell_z(\mathbf{Z}_m, \operatorname{sg}[\bar{\mathbf{Z}}_m]).
\label{eqn:lossfunction}
\end{align}
We use the MSE loss for $\ell_z(\mathbf{Z}_m, \bar{\mathbf{Z}}_m)$
and the cross-entropy loss for $\ell_y(\mathbf{Y}_m, \bar{\mathbf{Y}}_m)$. 
$\operatorname{sg}[\cdot]$ stands for stop gradient.
$\lambda$ is $2$ in our experiments.

\begin{figure*}
\centering
\includegraphics[width=.105\linewidth]{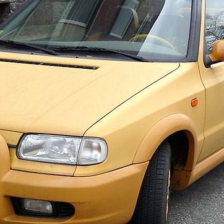}
\includegraphics[width=.105\linewidth]{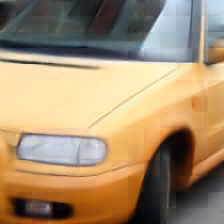}
\includegraphics[width=.105\linewidth]{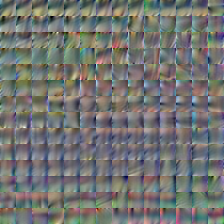}~
\includegraphics[width=.105\linewidth]{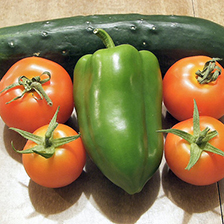}
\includegraphics[width=.105\linewidth]{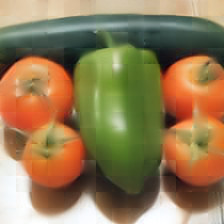}
\includegraphics[width=.105\linewidth]{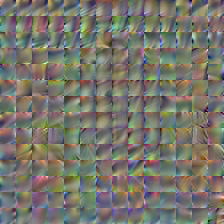}~
\includegraphics[width=.105\linewidth]{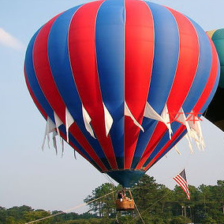}
\includegraphics[width=.105\linewidth]{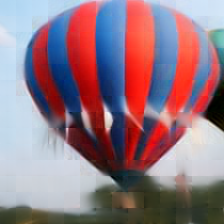}
\includegraphics[width=.105\linewidth]{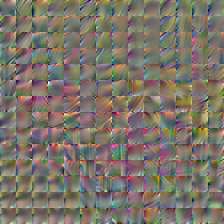}
\caption{
Illustrating that
predictions are made
in the representation space.
We reconstruct the image 
by feeding the full image 
($1$st, $4$th, and $7$th)  into the pretrained CAE encoder
and then the pretrained CAE decoder 
outputting the reconstructed image ($2$nd, $5$th, and $8$th).
It can be seen that
the image can be constructed
with the semantics kept
when skipping latent contextual regressor,
verifying the input and the predicted representations lie in the same space.
We also show the reconstructed images
($3$rd, $6$th, and $9$th)
from the encoder and the decoder
pretrained without the alignment constraint.
We can see that those images are meaningless,
indicating that the alignment constraint
is critical for ensuring 
that predictions are made
in the representation space.
}
\label{fig:nonmaskedimagereconstruction}
\end{figure*}

\section{Discussions}
\label{sec:discussionAnalysis}

\subsection{Analysis}
\noindent\textbf{Predictions
are made in the encoded representation space.}
Our CAE 
attempts to make predictions in the encoded representation space:
predict
the representations
for the masked patches
from the encoded representations of the visible patches.
In other words,
it is expected that
the output representations
of the latent contextual regressor
also lie in the encoded representation space,
which is ensured
by prediction alignment.
{This encourages the learned representation to take on a large extent of semantics for prediction
from visible patches
to masked patches, benefiting the representation learning of the encoder.}

We empirically verify
that the predicted representations
lie in the encoded representation space
through image reconstruction.
We train the CAE
using the pixel colors as the prediction targets,
for two cases:
with and without the alignment,
i.e., masked representation prediction.
For reconstruction, 
we feed all the patches (without masking,
all the image patches are visible)
of an image (from the ImageNet validation set)
into the pretrained encoder,
then skip the latent contextual regressor
and directly send all the encoded patch representations
to the pretrained decoder 
for reconstructing the whole image.

Figure~\ref{fig:nonmaskedimagereconstruction}
provides reconstruction results
for several examples randomly sampled 
from the ImageNet-$1$K validation set.
One can see that
our approach can successfully reconstruct the images,
implying that the input and  output representations of latent contextual regressor
are in the same space. 
In contrast,
without the alignment,
the reconstructed images are noisy,
indicating 
the input and output representations of latent contextual regressor
are in different spaces.
The results suggest that
the explicit prediction alignment
is critical for
ensuring that predictions are made 
in the encoded representation space. 

\vspace{1mm}
\noindent\textbf{Representation alignment in CAE and contrastive self-supervised learning.}
Representation alignment is also used in contrastive self-supervised learning methods,
such as MoCo, BYOL, SimCLR,
and methods mixing contrastive self-supervised learning and masked image modeling,
such as iBOT, and MST.
The alignment loss could be the MSE loss or 
the contrastive loss that CAE may also take advantage of.

In the CAE, the alignment is imposed
over the representations
$\mathbf{Z}_m = \mathcal{H}(\mathcal{F}(\mathbf{X}_v))$ - 
predicted from the representations $\mathcal{F}(\mathbf{X}_v)$
of visible patches through the regressor $\mathcal{H}$,
and the representations 
$\bar{\mathbf{Z}}_m
=\mathcal{F}(\mathbf{X}_m)$ 
- computed from the encoder $\mathcal{F}$.
Both \emph{$\mathbf{Z}_m$ and $\bar{\mathbf{Z}}_m$ are about the masked patches,
and lie in the representation space
output from the encoder}.

Differently,
the alignment in the  most 
contrastive self-supervised learning methods
is imposed 
over the representations $\{\mathcal{P}(\mathcal{F}(\mathbf{V}_1)),
\mathcal{P}(\mathcal{F}(\mathbf{V}_2)),
\cdots,
\mathcal{P}(\mathcal{F}((\mathbf{V}_N))\}$,
where $\mathcal{P}$ is a projector,
and some views may be processed
with the EMA version of the encoder
and the projector.
The $N$ representations to be aligned
are about \emph{different views} $\{\mathbf{V}_1,
\mathbf{V}_2,\cdots, \mathbf{V}_N\}$
(in iBoT and MST,
the views are masked views and full views),
and are not directly output from the encoder.
It is not quite clear how the projector works,
and it is reported in~\cite{MIMPart2022}
that the projector is a part-to-whole process
mapping the object part representation
to the whole object representation
for contrastive self-supervised learning.

\subsection{Connection}
\noindent\textbf{Relation to autoencoder.}
The original autoencoder~\cite{phdthesis_LeCun,gallinari1987memoires,hinton1994autoencoders} consists of
an encoder and a decoder.
The encoder maps the input into a latent representation,
and the decoder reconstructs the input
from the latent representation.
The denoising autoencoder (DAE)~\cite{VincentLLBM10},
a variant of autoencoder,
corrupts the input by adding noises
and still reconstructs
the non-corrupted input.

Our CAE encoder is similar
to the original autoencoder
and also contains an encoder and a decoder.
Different from the autoencoder
where the encoder and the decoder
process the whole image,
our encoder takes a portion of patches as input
and our decoder takes
the estimated latent representations
of the other portion of patches
as input.
Importantly,
the CAE makes predictions
in the latent space
from the visible patches
to the masked patches.

\definecolor{cornflowerblue}{RGB}{100, 149, 237}
\begin{figure*}[t]
\vskip 0.1in
\centering
\footnotesize
\subfigure[]{\fbox{\includegraphics[scale=0.55]{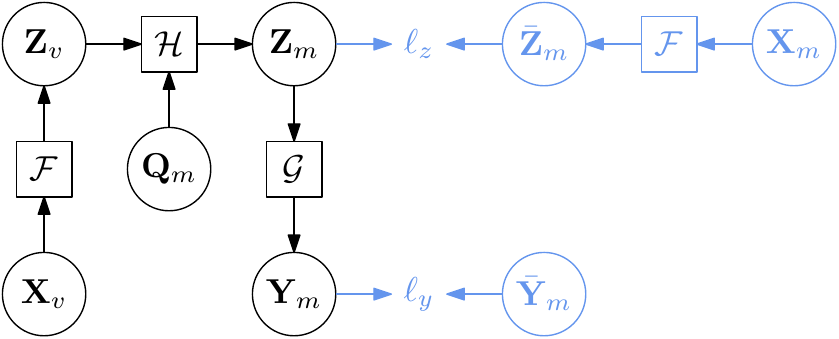}}}~~~
\subfigure[]{\fbox{\includegraphics[scale=0.55]{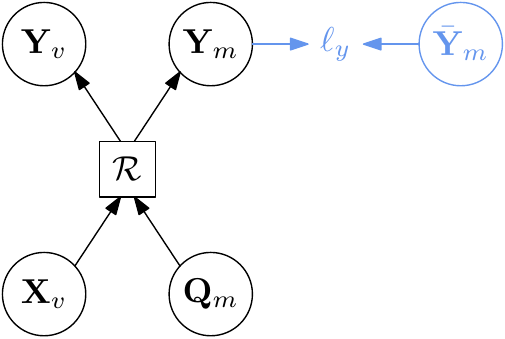}}}~~~
\subfigure[]{\fbox{\includegraphics[scale=0.55]{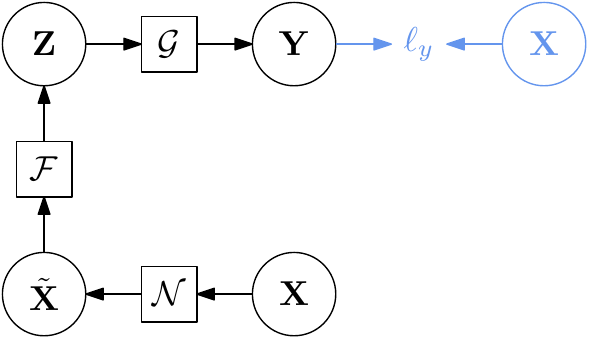}}}~~~
\subfigure[]{\fbox{\includegraphics[scale=0.55]{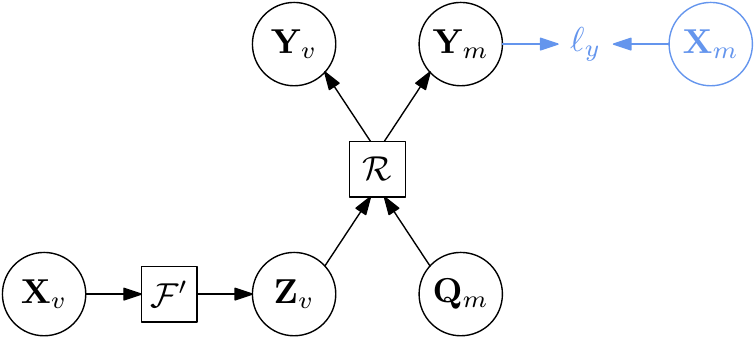}}}
\caption{The computational graphs
for (a) a context autoencoder (CAE),
(b) BEiT~\cite{bao2021beit},
(c) a denoising autoencoder (DAE),
and (d) MAE~\cite{he2021masked}
and the one stream in SplitMask~\cite{el2021large}.
The parts in {\color{cornflowerblue}cornflower blue}
are for loss function.
(a) 
The encoder $\mathcal{F}$
receives visible patches
$\mathbf{X}_v$
and outputs their latent representations 
$\mathbf{Z}_v$.
The latent contextual regressor
$\mathcal{H}$
predicts the latent representations 
$\mathbf{Z}_m$
for masked patches 
from $\mathbf{Z}_v$.
The decoder predicts the targets $\mathbf{Y}_m$
for masked patches
from $\mathbf{Z}_m$.
$\ell_z$ and $\ell_y$
are the loss functions.
During training, the gradient is stopped for $\bar{\mathbf{Z}}_m$.
See the detail in Section~\ref{sec:cae}.
(b) 
The input includes both visible patches
$\mathbf{X}_v$
and mask queries $\mathbf{Q}_m$ representing masked patches,
and the representations for them are updated within the function $\mathcal{R}$.
(c) The function $\mathcal{N}$ is a noising function
generating the noisy version $\tilde{\mathbf{X}}$
from the input $\mathbf{X}$.
$\mathcal{F}$ and $\mathcal{G}$ are the normal encoder and decoder, respectively.
(d) The two functions, $\mathcal{F}'$ 
and $\mathcal{R}$,
are both based on self-attention.
$\mathcal{F}'$ 
(called encoder in MAE)
only processes the visible patches $\mathbf{X}_v$,
and $\mathcal{R}$
(called decoder in MAE)
processes 
both the latent representations $\mathbf{Z}_v$
of the visible patches
and the mask queries ($\mathbf{Q}_m$)
and updates them simultaneously.
For simplicity,
the positional embeddings 
are not included in computational graphs.
\emph{
(a) CAE and (c) DAE 
perform
the encoding and  MIM task completion roles explicitly
and separately,
(b) BEiT and (d) MAE perform the encoding and MIM task completion roles implicitly
and simultaneously.}
}
\label{fig:ComputationGraph}
\end{figure*}

\begin{figure*}[t]
\centering
\footnotesize
\includegraphics[width=.1\linewidth]{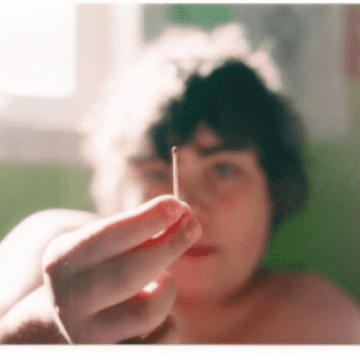}~
\includegraphics[width=.1\linewidth]{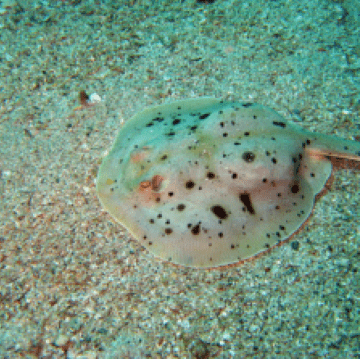}~
\includegraphics[width=.1\linewidth]{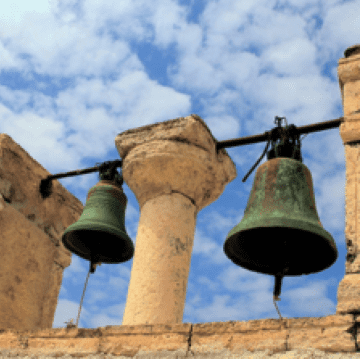}~
\includegraphics[width=.1\linewidth]{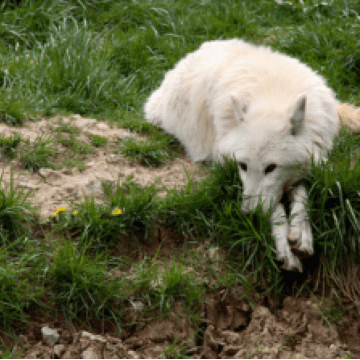}~
\includegraphics[width=.1\linewidth]{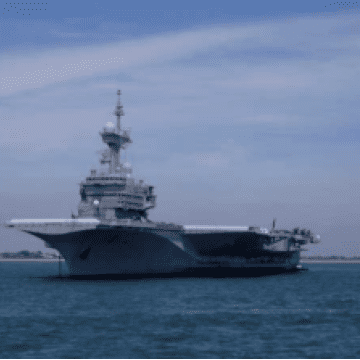}~
\includegraphics[width=.1\linewidth]{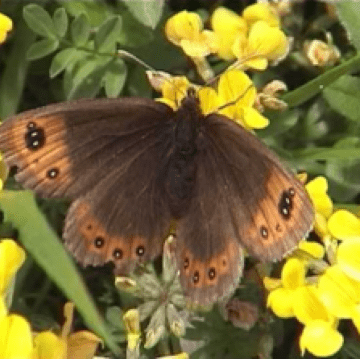}~
\includegraphics[width=.1\linewidth]{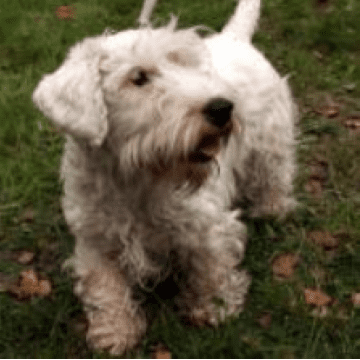}~
\includegraphics[width=.1\linewidth]{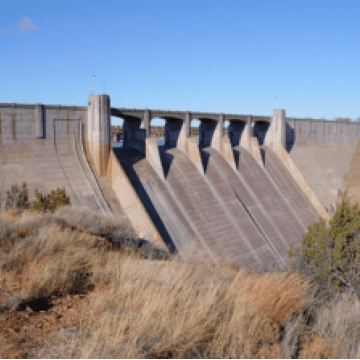}~
\includegraphics[width=.1\linewidth]{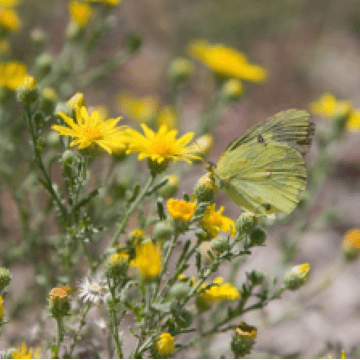}\\
\vspace{1mm}
\includegraphics[width=.1\linewidth]{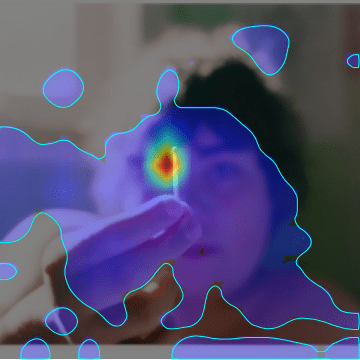}~
\includegraphics[width=.1\linewidth]{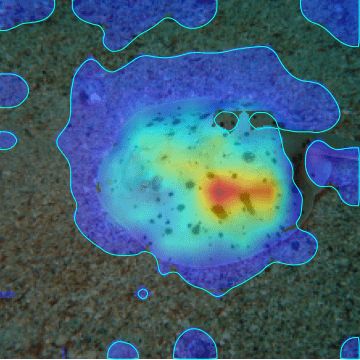}~
\includegraphics[width=.1\linewidth]{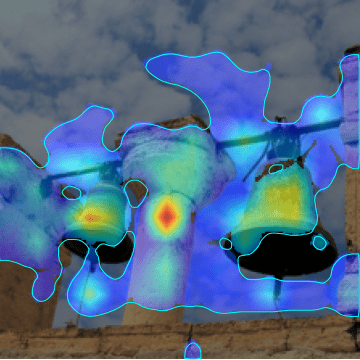}~
\includegraphics[width=.1\linewidth]{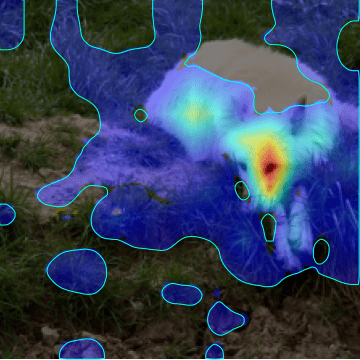}~
\includegraphics[width=.1\linewidth]{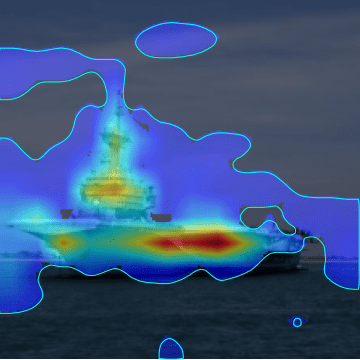}~
\includegraphics[width=.1\linewidth]{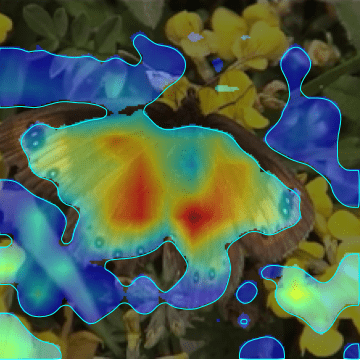}~
\includegraphics[width=.1\linewidth]{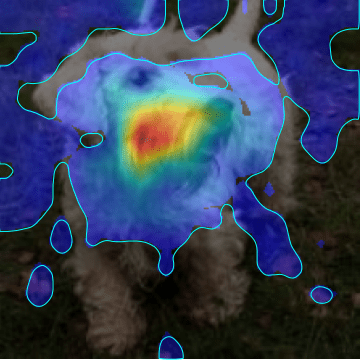}~
\includegraphics[width=.1\linewidth]{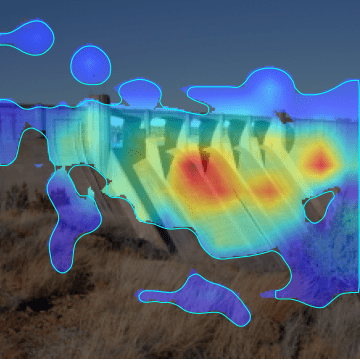}~
\includegraphics[width=.1\linewidth]{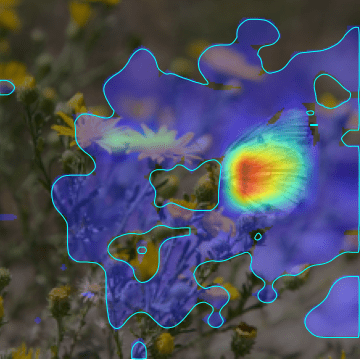}\\
\vspace{1mm}
\includegraphics[width=.1\linewidth]{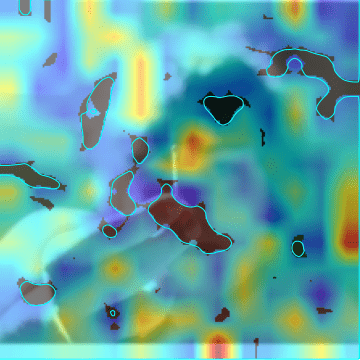}~
\includegraphics[width=.1\linewidth]{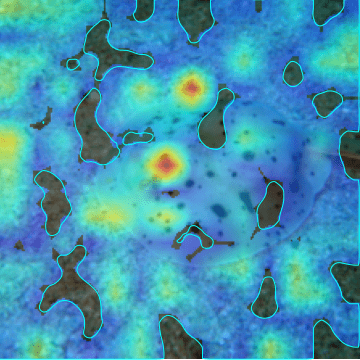}~
\includegraphics[width=.1\linewidth]{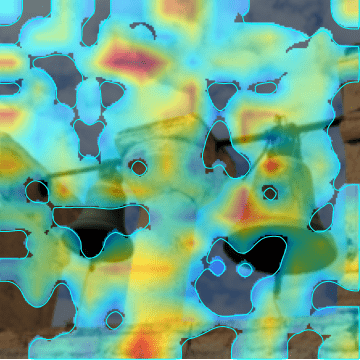}~
\includegraphics[width=.1\linewidth]{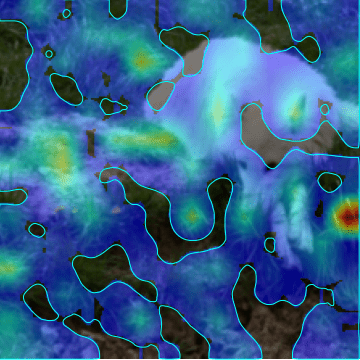}~
\includegraphics[width=.1\linewidth]{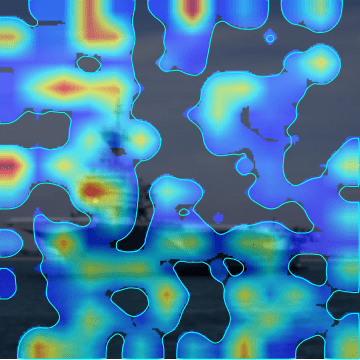}~
\includegraphics[width=.1\linewidth]{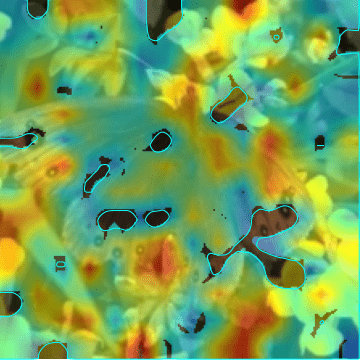}~
\includegraphics[width=.1\linewidth]{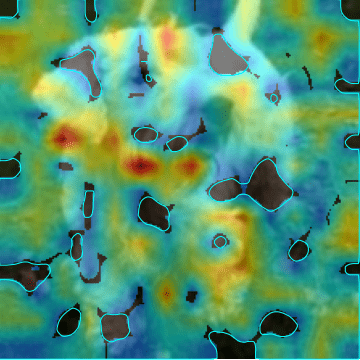}~
\includegraphics[width=.1\linewidth]{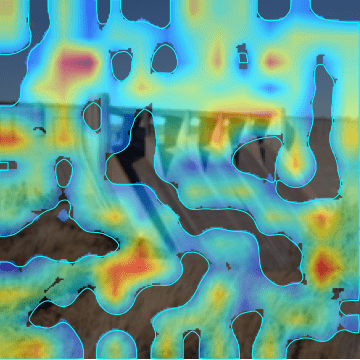}~
\includegraphics[width=.1\linewidth]{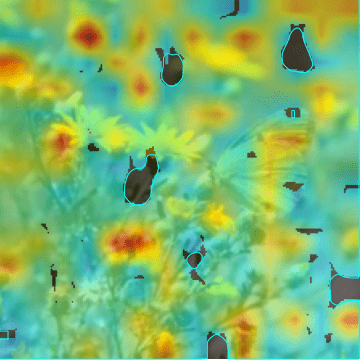}\\
\vspace{1mm}
\includegraphics[width=.1\linewidth]{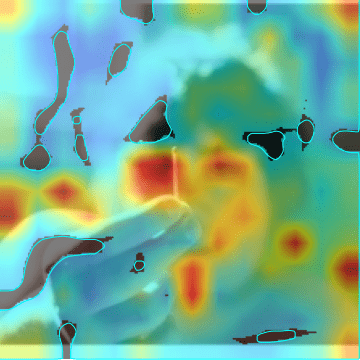}~
\includegraphics[width=.1\linewidth]{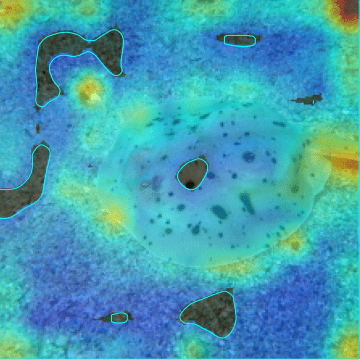}~
\includegraphics[width=.1\linewidth]{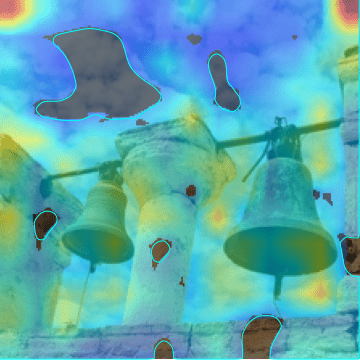}~
\includegraphics[width=.1\linewidth]{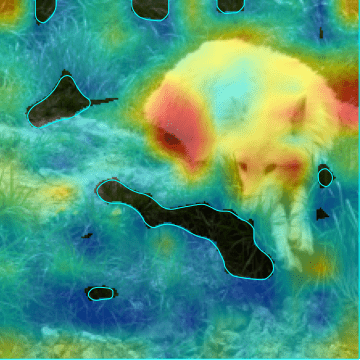}~
\includegraphics[width=.1\linewidth]{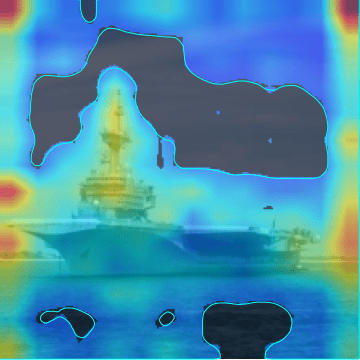}~
\includegraphics[width=.1\linewidth]{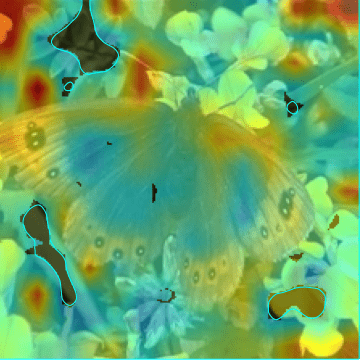}~
\includegraphics[width=.1\linewidth]{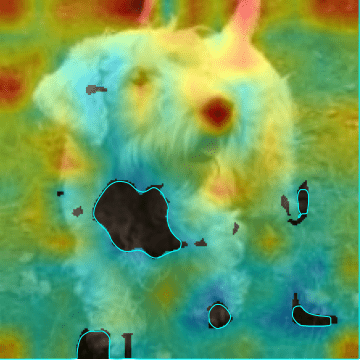}~
\includegraphics[width=.1\linewidth]{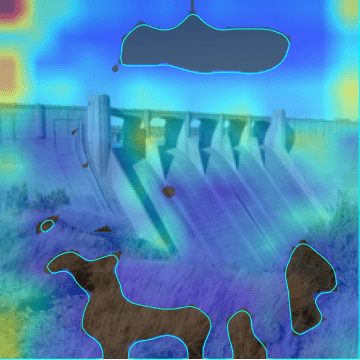}~
\includegraphics[width=.1\linewidth]{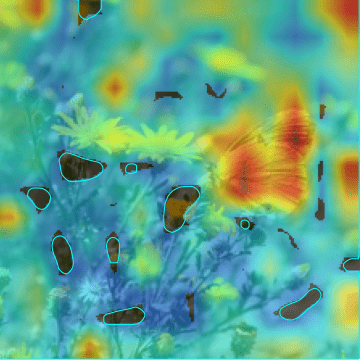}\\
\caption{Illustrating
the attention map
averaged over $12$ attention heads
between the class token
and the patch tokens 
in the last layer
of the ViT encoder
pretrained on ImageNet-$1$K.
The region inside the blue contour is obtained by thresholding the attention weights to keep $50\%$ of the mass.
The four rows are: (1) input image,
(2) MoCo v3,
a typical contrastive self-supervised learning method,
(3) MAE, and (4) our CAE.
One can see that 
MoCo v3 tends to focus 
mainly on the centering regions 
and little on other patches,
and our CAE tends to consider
almost all the patches.
}
\label{fig:patchimportance}
\vspace{-0.4cm}
\end{figure*}

\vspace{1mm}
\noindent\textbf{Relation to BEiT, iBoT and MAE.}
The CAE encoder processes
the visible patches,
to extract their representations,
without making predictions for masked patches.
Masked representation prediction 
is made through the regressor and the prediction alignment,
ensuring that the output of the regressor
lies in the representation space same with 
the encoder output.
The decoder only processes
the predicted representations
of masked patches.
Our approach encourages that 
the encoder takes the responsibility
of 
and is only for 
representation learning.

In contrast,
BEiT~\cite{bao2021beit}
and the MIM part of iBOT do not separate 
the representation extraction role
and the task completion role
and uses a single network,
with both the visible and masked patches
as the input,
simultaneously for the two roles.
In MAE~\cite{he2021masked}, the so-called decoder
may play a partial role
for representation learning
as the representations of the visible patches
are also updated in the MAE decoder. 
Unlike CAE, MAE, iBoT, BEiT
do not explicitly 
predict the representations
of masked patches from
the representations of visible patches
(that lie in the encoded representation space)
for masked patches.

When the pretrained encoder is applied to downstream tasks,
one often replaces the pretext task completion part
using the downstream task layer,
e.g., segmentation layer or detection layer.
The separation of representation learning (encoding)
and pretext task completion
helps that downstream task applications
take good advantage of representation pretraining.

We provide 
the computational graph 
for CAE, BEiT~\cite{bao2021beit}, denoising autoencoder, Masked Autoencoder~\cite{he2021masked}
and SplitMask~\cite{el2021large}
(one stream)
in Figure~\ref{fig:ComputationGraph}.
Compared to our CAE,
the main issue of MAE is that the so-called decoder $\mathcal{R}$
might have also the encoding role,
i.e.,
learning semantic representations of
the visible patches.

 \vspace{1mm}
\noindent\textbf{Comparison to contrastive self-supervised learning.}
Typical contrastive self-supervised learning methods,
e.g., SimCLR~\cite{ChenK0H20}
and MoCo~\cite{He0WXG20,ChenXH21},
pretrain the networks
by solving the pretext task,
maximizing the similarities between augmented views
(e.g., random crops)
from the same image
and minimizing the similarities
between augmented views from different images.

It is shown in~\cite{ChenK0H20} that
random cropping
plays an important role in view augmentation 
for contrastive self-supervised learning.
Through analyzing random crops (illustrated in Figure~\ref{fig:maskingandcropping}),
we observe that
the center pixels 
in the original image space
have large chances to belong to random crops.
We suspect that
the global representation,
learned by contrastive self-supervised learning
for a random crop 
possibly
with other augmentation schemes,
tends to 
focus mainly on the center pixels
in the original image,
so that the representations
of different crops 
from the same image
can be possibly similar.
Figure~\ref{fig:patchimportance}
(the second row)
shows that 
the center region of the original image
for the typical contrastive self-supervised learning approach, MoCo v3,
is highly attended.
The part in random crops corresponding
to the center of the original image
is still attended
as shown in Figure~\ref{fig:patchimportance_crop}.

In contrast,
our CAE method
(and other MIM methods)
randomly samples the patches
from the augmented views
to form the visible and masked patches.
All the patches are possible
to be randomly masked
for the augmented views and accordingly the original image.
Thus, the CAE encoder needs to learn good representations
for all the patches,
to make good predictions
for the masked patches
from the visible patches.
Figure~\ref{fig:patchimportance} (the third row)
illustrates that almost all the patches in the original images
are considered in our CAE encoder.

Considering that
the instances of the $1000$ categories
in ImageNet-$1$K
locate mainly around the center of the original images~\cite{russakovsky2015imagenet},
typical contrastive self-supervised learning methods, e.g., MoCo v3,
learn the knowledge mainly about the $1000$ categories,
which is similar to supervised pretraining.
But our CAE and other MIM methods
are able to  
learn more knowledge beyond
the $1000$ categories
from the non-center image regions.
This indicates that
the CAE has the potential to perform better
for downstream tasks.

\begin{figure}[t]
\centering

\setlength{\fboxsep}{1pt}
\setlength{\fboxrule}{0.4pt}
\fbox{\includegraphics[width=0.45\columnwidth]{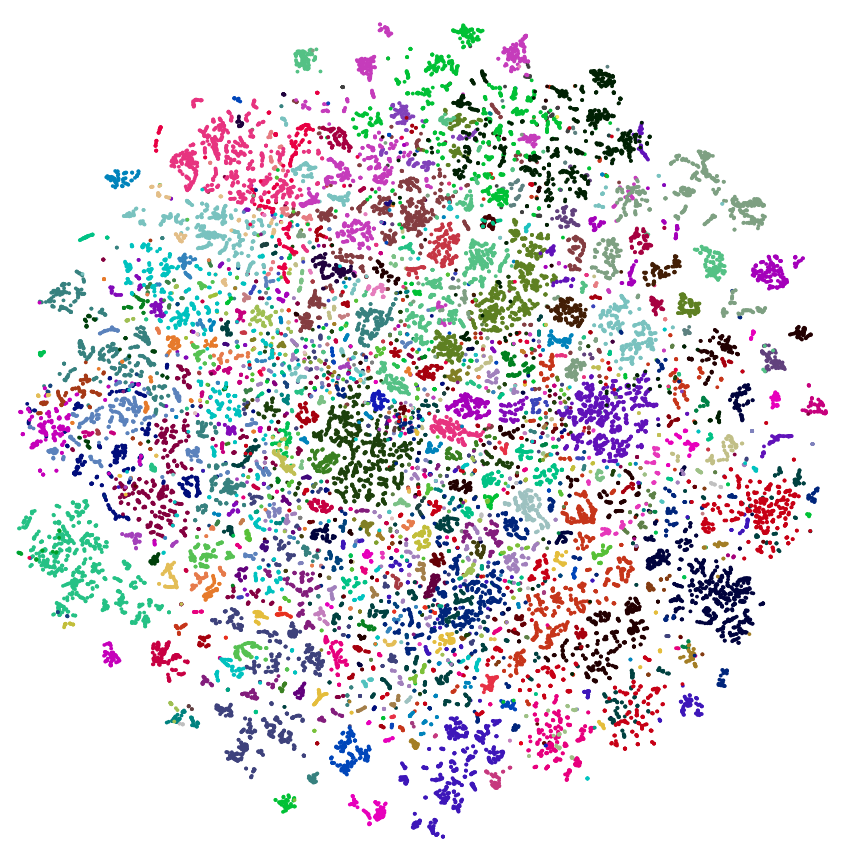}}~~~~
\fbox{\includegraphics[width=0.45\columnwidth]{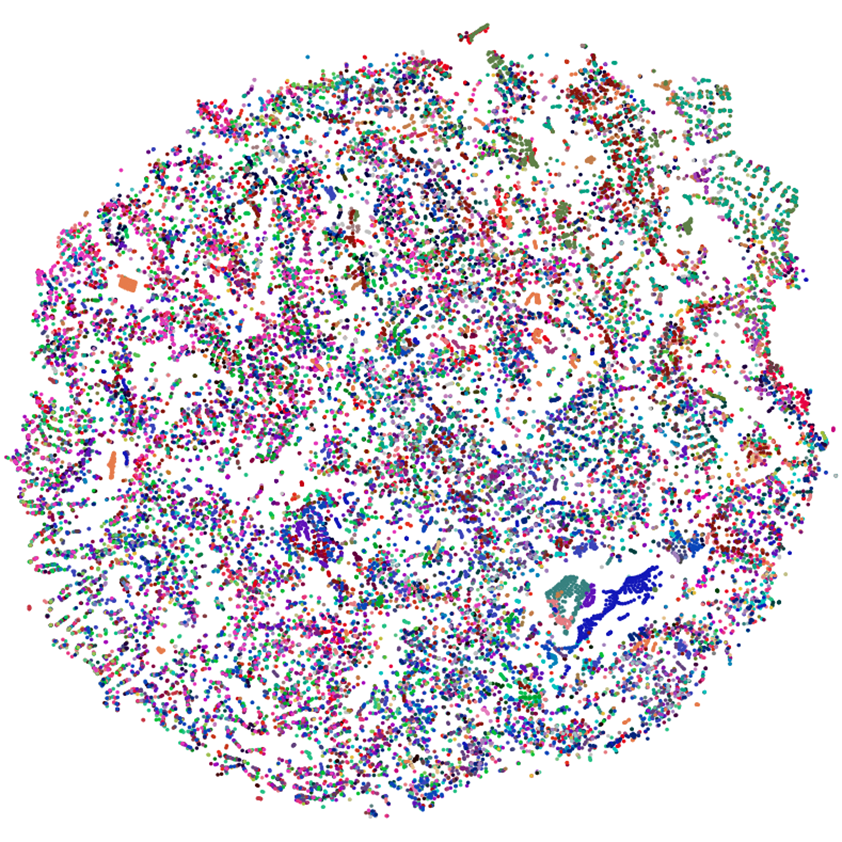}}
\caption{t-SNE visualization
(one color for one category)
of representations extracted
from the images in ADE$20$K. Left: ViT pretrained with our CAE; Right: ViT with random weights.
}
\label{fig:representationclusters}
\end{figure}

\subsection{Interpretation}
\noindent\textbf{Intuitive Interpretation for CAE.}
Humans are able to
hallucinate 
what appears in the masked regions
and how they appear
according to the visible regions.
We speculate that
humans do this
possibly 
in a way similar as
the following example:
given that only the region of the dog's head
is visible and the remaining parts are missing,
one can (a) recognize the visible region 
to be about a dog,
(b) predict the regions 
where the other parts of the dog appear,
and (c) guess what the other parts look like.

Our CAE encoder is in some sense
like the human recognition step (a).
It understands the content
by
mapping the visual patches
into latent representations
that lie in the subspace that corresponds to the category dog\footnote{Our encoder does not know that the subspace is about a dog, and just separates it from the subspaces of other categories.}.
The latent contextual regressor
is like step (b).
It produces a plausible hypothesis
for the masked patches,
and describes
the regions corresponding to the other parts of the dog
using latent representations.
The CAE decoder is like step (c),
mapping the latent representations
to the targets.
It should be noted that
the latent representations 
might contain other information
besides the semantic information,
e.g., the part information
and the information for making predictions.

We adopt t-SNE~\cite{van2008visualizing}
to visualize the high-dimensional patch representations 
output from our CAE encoder 
on ADE$20$K~\cite{zhou2017scene} in Figure~\ref{fig:representationclusters}.
ADE$20$K has a total of $150$ categories. 
For each patch in the image, 
we set its label to be the category
that more than half of the pixels
belong to.
We collect up to $1000$ patches for each category
from sampled $500$ images. As shown in the figure, the latent representations of CAE are clustered to some degree for different categories (though not perfect as our CAE is pretrained on ImageNet-1K).
Similar observations could be found for other MIM methods.

\begin{figure*}[t]
\centering
\footnotesize
\includegraphics[width=.1\linewidth]{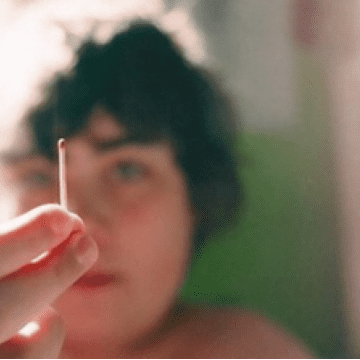}~~
\includegraphics[width=.1\linewidth]{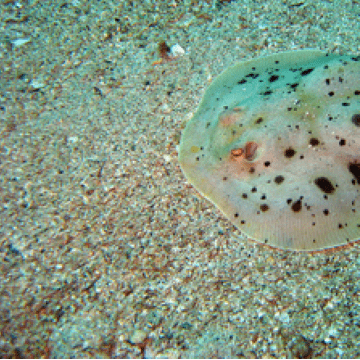}~~
\includegraphics[width=.1\linewidth]{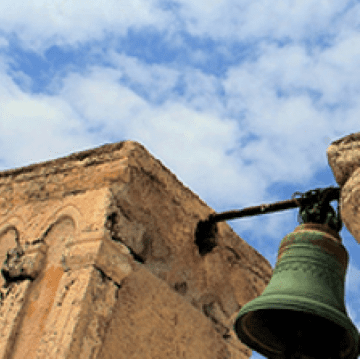}~~
\includegraphics[width=.1\linewidth]{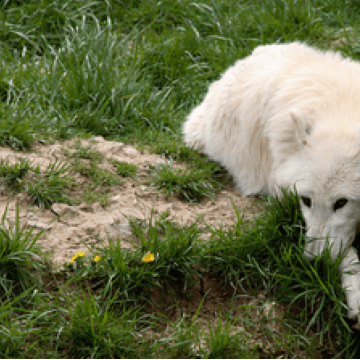}~~
\includegraphics[width=.1\linewidth]{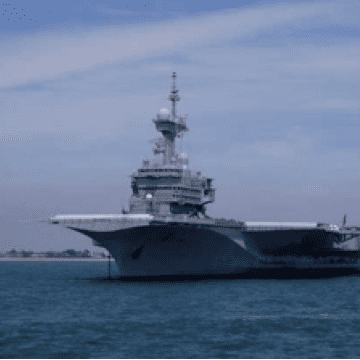}~~
\includegraphics[width=.1\linewidth]{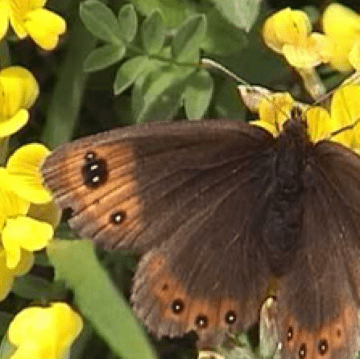}~~
\includegraphics[width=.1\linewidth]{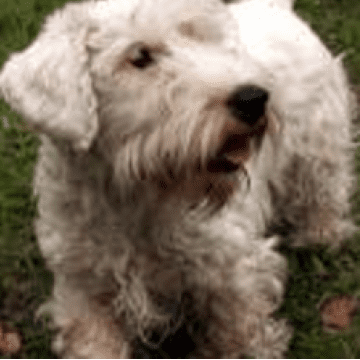}~~
\includegraphics[width=.1\linewidth]{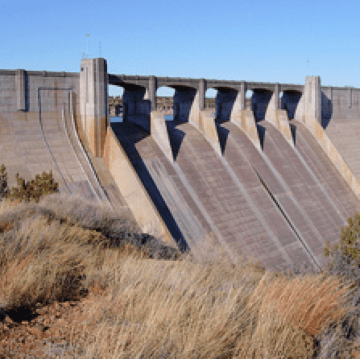}~~
\includegraphics[width=.1\linewidth]{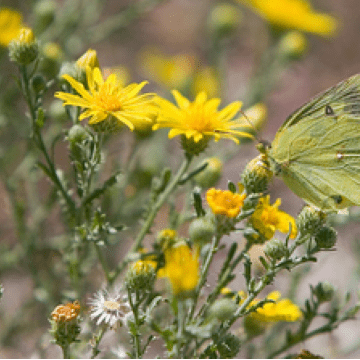}~~
\\
\vspace{1mm}
\includegraphics[width=.1\linewidth]{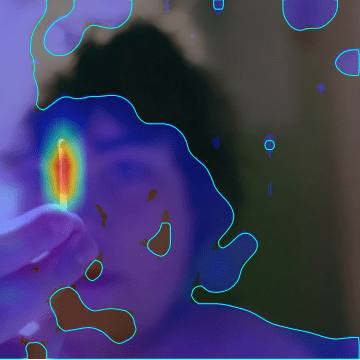}~~
\includegraphics[width=.1\linewidth]{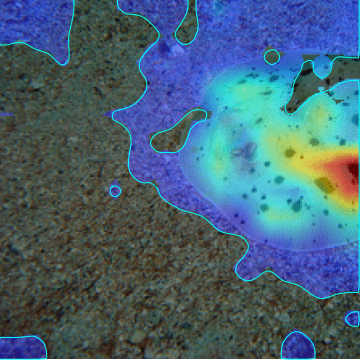}~~
\includegraphics[width=.1\linewidth]{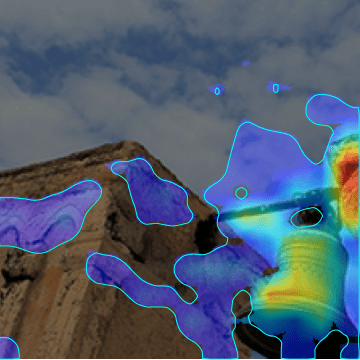}~~
\includegraphics[width=.1\linewidth]{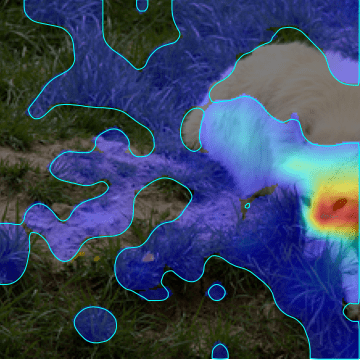}~~
\includegraphics[width=.1\linewidth]{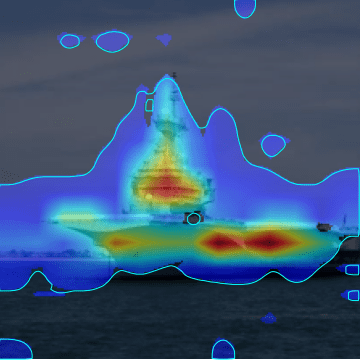}~~
\includegraphics[width=.1\linewidth]{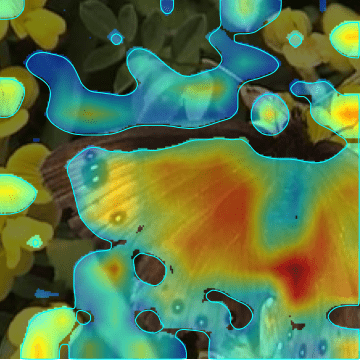}~~
\includegraphics[width=.1\linewidth]{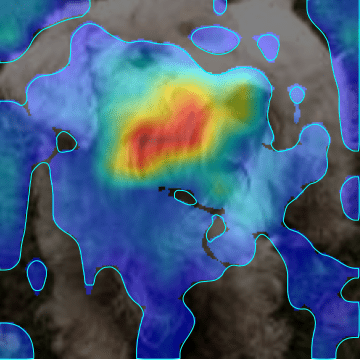}~~
\includegraphics[width=.1\linewidth]{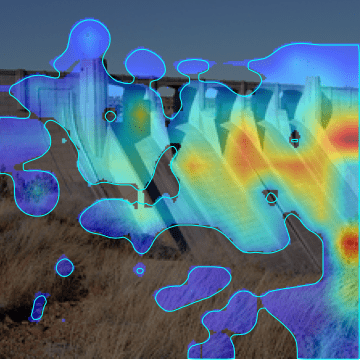}~~
\includegraphics[width=.1\linewidth]{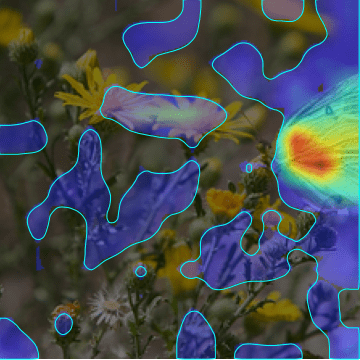}~~
\\
\vspace{1mm}
\includegraphics[width=.1\linewidth]{mae_253_scope.png}~~
\includegraphics[width=.1\linewidth]{mae_273_scope.png}~~
\includegraphics[width=.1\linewidth]{mae_322_scope.png}~~
\includegraphics[width=.1\linewidth]{mae_315_scope.png}~~
\includegraphics[width=.1\linewidth]{mae_137_scope.png}~~
\includegraphics[width=.1\linewidth]{mae_22_scope.png}~~
\includegraphics[width=.1\linewidth]{mae_161_scope.png}~~
\includegraphics[width=.1\linewidth]{mae_24_scope.png}~~
\includegraphics[width=.1\linewidth]{mae_92_scope.png}~~
\\
\vspace{1mm}
\includegraphics[width=.1\linewidth]{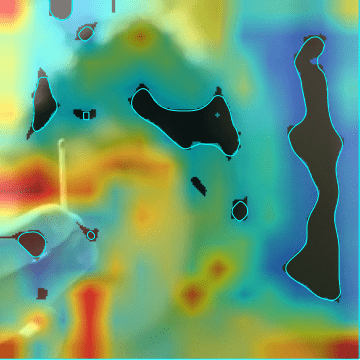}~~
\includegraphics[width=.1\linewidth]{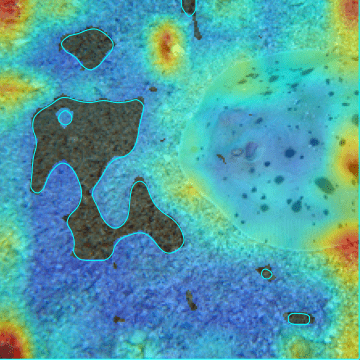}~~
\includegraphics[width=.1\linewidth]{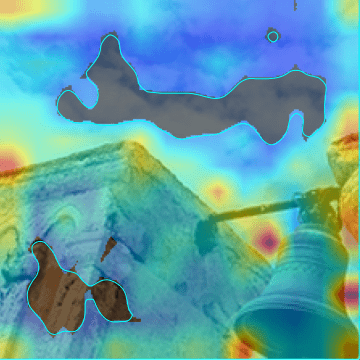}~~
\includegraphics[width=.1\linewidth]{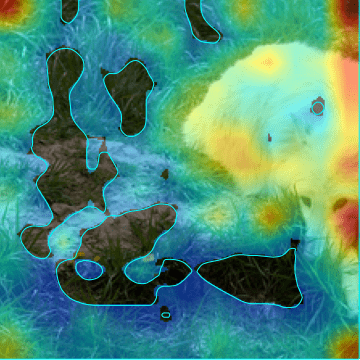}~~
\includegraphics[width=.1\linewidth]{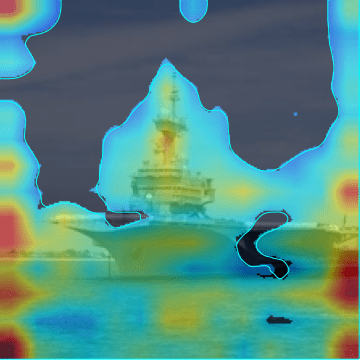}~~
\includegraphics[width=.1\linewidth]{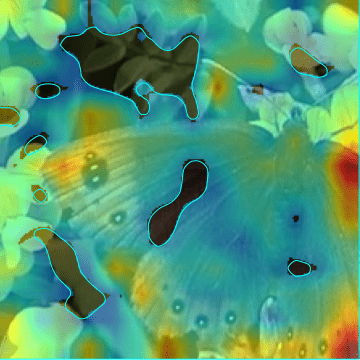}~~
\includegraphics[width=.1\linewidth]{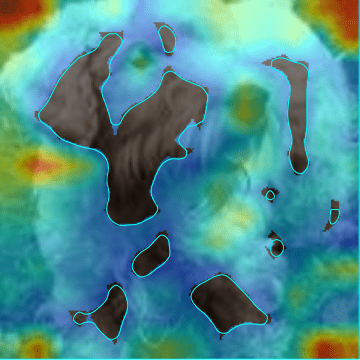}~~
\includegraphics[width=.1\linewidth]{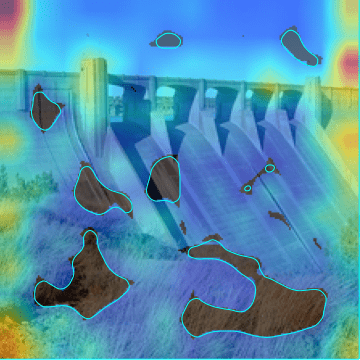}~~
\includegraphics[width=.1\linewidth]{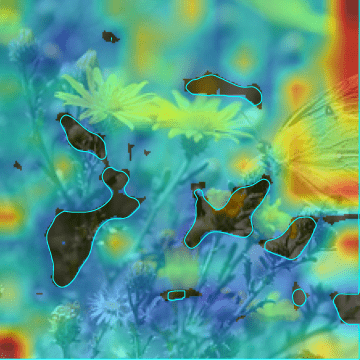}~~
\\
\vspace{1mm}
\includegraphics[width=.1\linewidth]{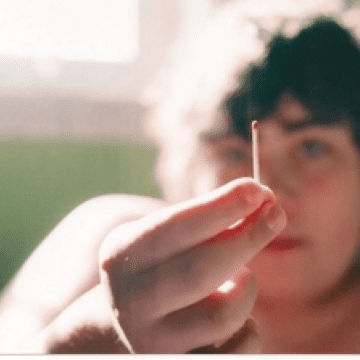}~~
\includegraphics[width=.1\linewidth]{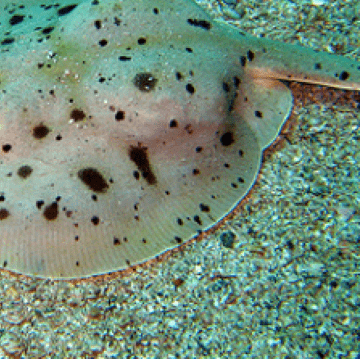}~~
\includegraphics[width=.1\linewidth]{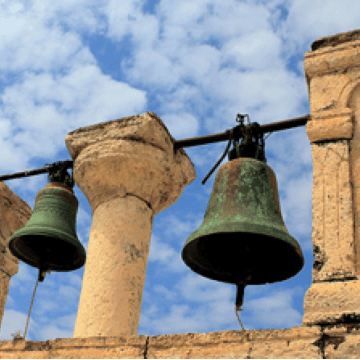}~~
\includegraphics[width=.1\linewidth]{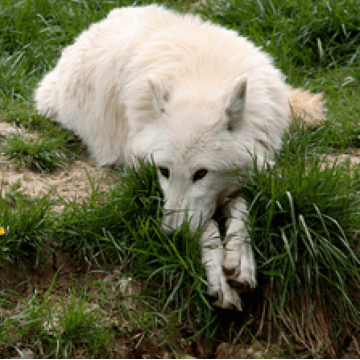}~~
\includegraphics[width=.1\linewidth]{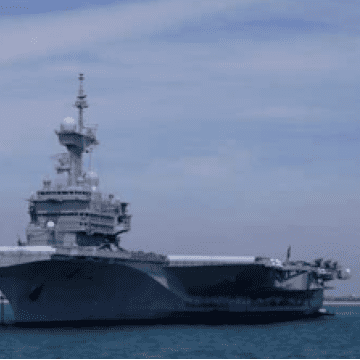}~~
\includegraphics[width=.1\linewidth]{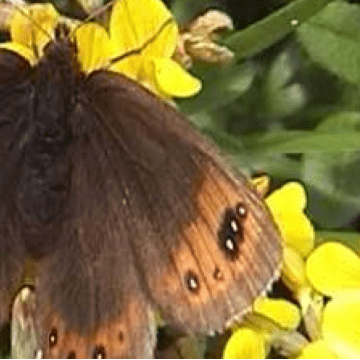}~~
\includegraphics[width=.1\linewidth]{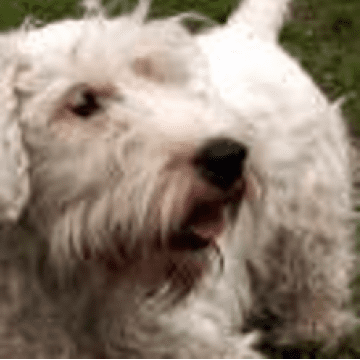}~~
\includegraphics[width=.1\linewidth]{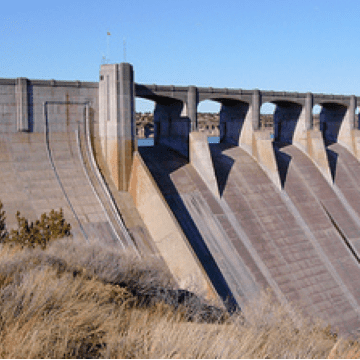}~~
\includegraphics[width=.1\linewidth]{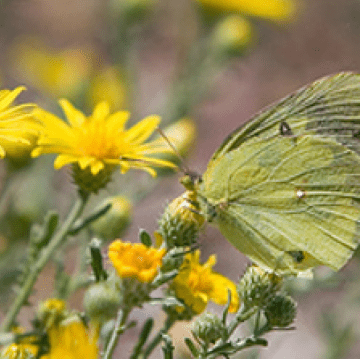}~~
\\
\vspace{1mm}
\includegraphics[width=.1\linewidth]{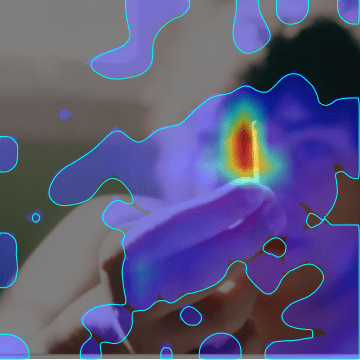}~~
\includegraphics[width=.1\linewidth]{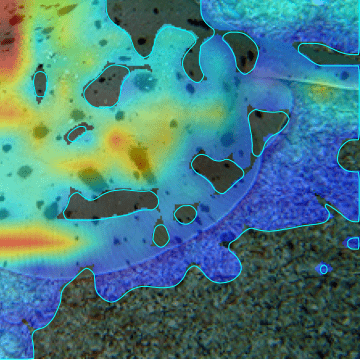}~~
\includegraphics[width=.1\linewidth]{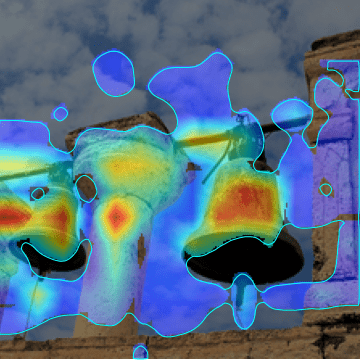}~~
\includegraphics[width=.1\linewidth]{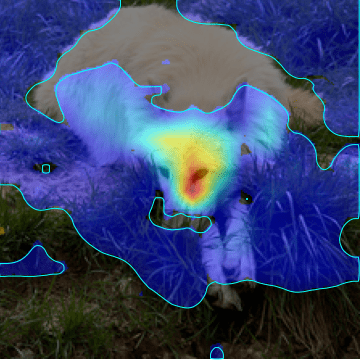}~~
\includegraphics[width=.1\linewidth]{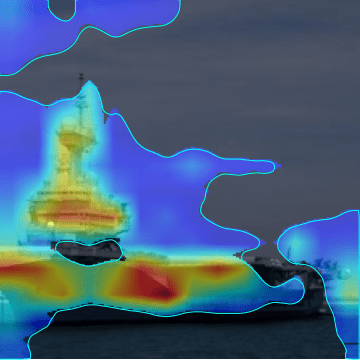}~~
\includegraphics[width=.1\linewidth]{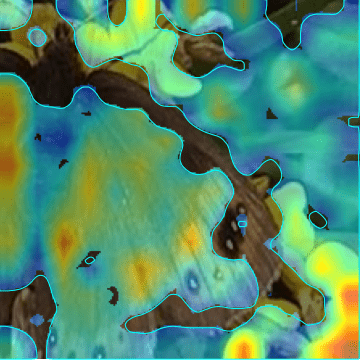}~~
\includegraphics[width=.1\linewidth]{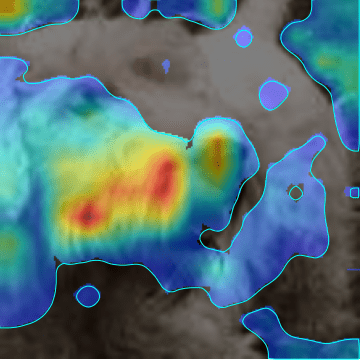}~~
\includegraphics[width=.1\linewidth]{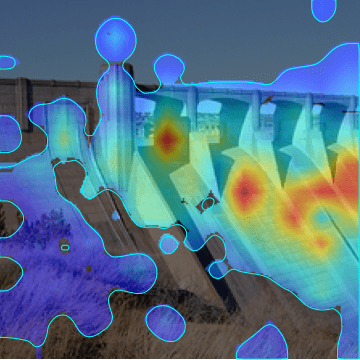}~~
\includegraphics[width=.1\linewidth]{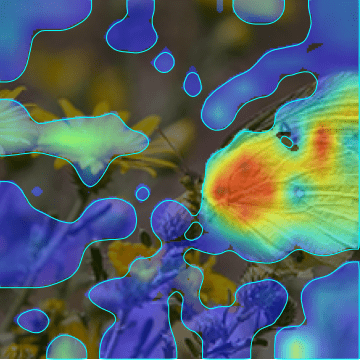}~~
\\
\vspace{1mm}
\includegraphics[width=.1\linewidth]{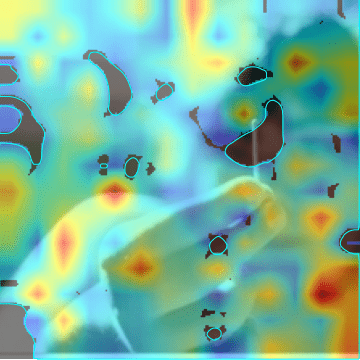}~~
\includegraphics[width=.1\linewidth]{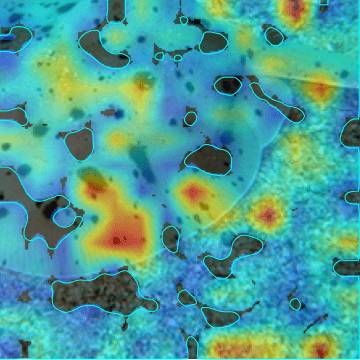}~~
\includegraphics[width=.1\linewidth]{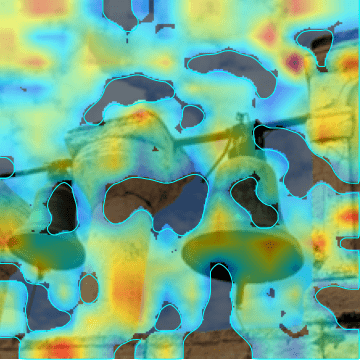}~~
\includegraphics[width=.1\linewidth]{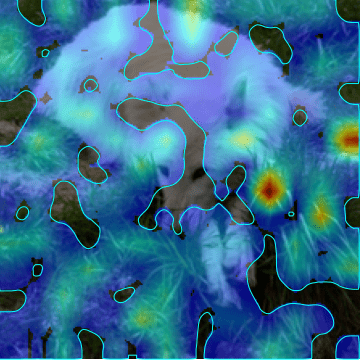}~~
\includegraphics[width=.1\linewidth]{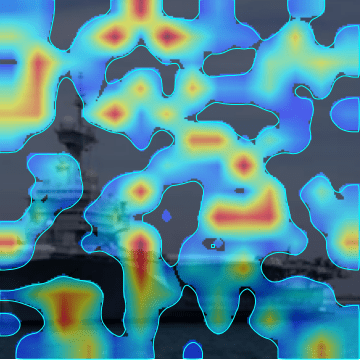}~~
\includegraphics[width=.1\linewidth]{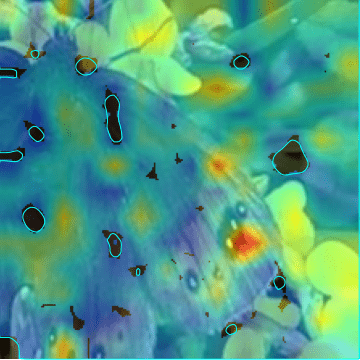}~~
\includegraphics[width=.1\linewidth]{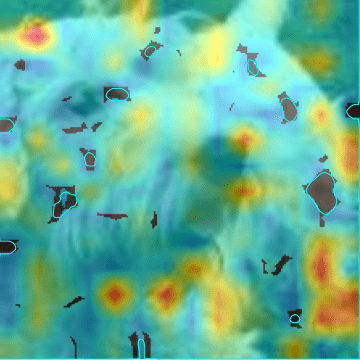}~~
\includegraphics[width=.1\linewidth]{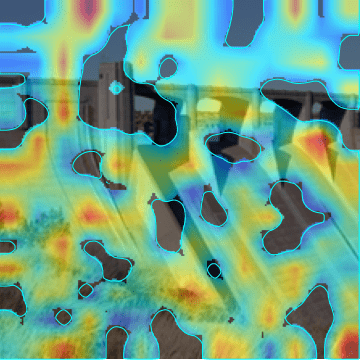}~~
\includegraphics[width=.1\linewidth]{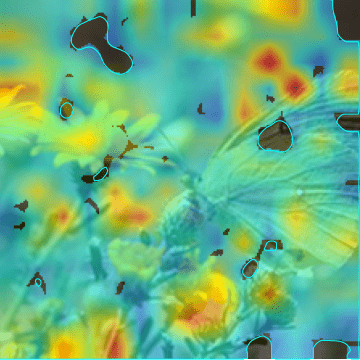}~~
\\
\vspace{1mm}
\includegraphics[width=.1\linewidth]{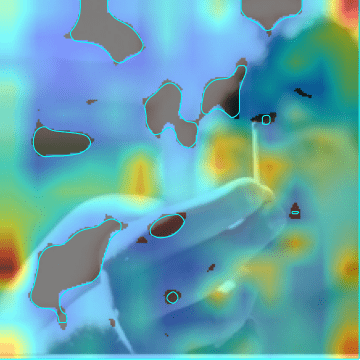}~~
\includegraphics[width=.1\linewidth]{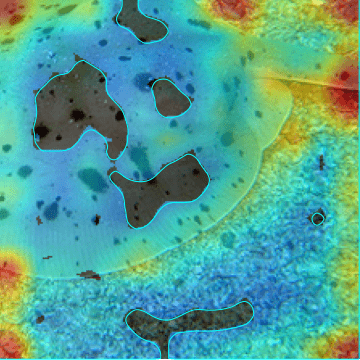}~~
\includegraphics[width=.1\linewidth]{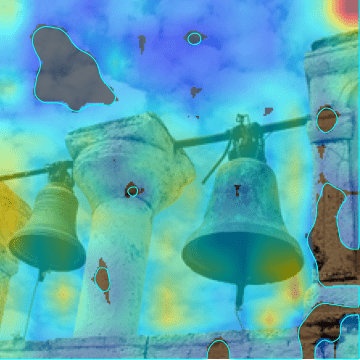}~~
\includegraphics[width=.1\linewidth]{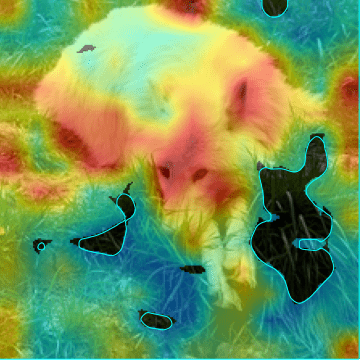}~~
\includegraphics[width=.1\linewidth]{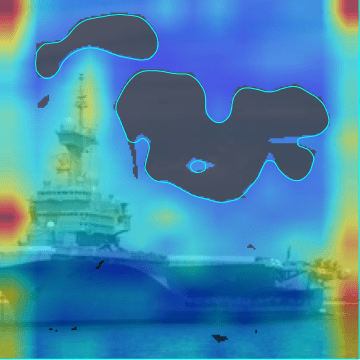}~~
\includegraphics[width=.1\linewidth]{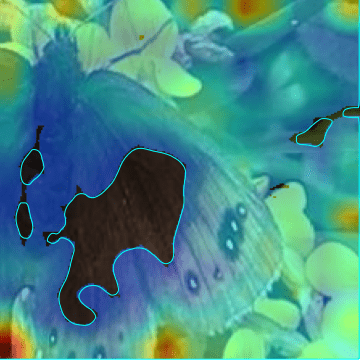}~~
\includegraphics[width=.1\linewidth]{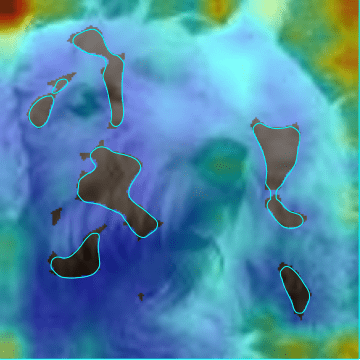}~~
\includegraphics[width=.1\linewidth]{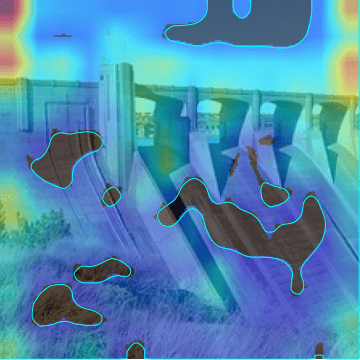}~~
\includegraphics[width=.1\linewidth]{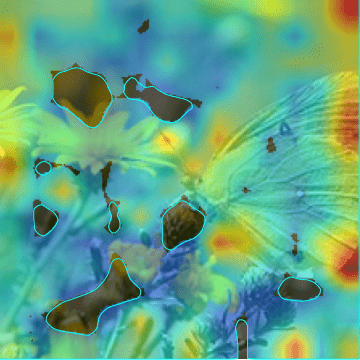}~~
\\
\caption{The attention maps over two sets of randomly cropped images (the $1$st the $5$th rows)
for MoCo v3 (the $2$nd the $6$th rows), MAE (the $3$rd the $7$th rows), and our CAE (the $4$th the $8$th rows)
pretrained on ImageNet-$1$K.
The contrastive self-supervised learning method, MoCo v3, tends to focus 
mainly on the object region 
and little on other regions.
In contrast, MIM-based models, CAE and MAE, tend to consider
almost all the patches.
The attention maps over the original images
are shown in Figure~\ref{fig:patchimportance}.}
\label{fig:patchimportance_crop}
\end{figure*}

\begin{figure}
\centering
\footnotesize
\includegraphics[scale=0.8]{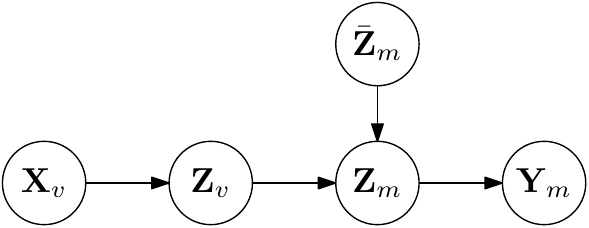}
\caption{The probabilistic graphical model of CAE.
The other conditions
of $\mathbf{Z}_v$,
$\mathbf{Z}_m$,
and $\mathbf{Y}_m$,
the positions $\mathbf{P}_v$
and $\mathbf{P}_m$ 
of the visible and masked patches,
are not plotted for simplicity.}
\label{fig:CAEPGM}
\end{figure}

\vspace{1mm}
\noindent\textbf{Probabilistic interpretation for CAE.}
The MIM problem can be formulated 
in the probabilistic form, maximizing
the probability of the predictions $\mathbf{Y}_m$ 
of the masked patches
given the conditions, the visible patches $\mathbf{X}_v$,
the positions $\mathbf{P}_v$ of the visible patches,
and the positions $\mathbf{P}_m$ of the masked patches:
$P(\mathbf{Y}_m \mid \mathbf{X}_v, \mathbf{P}_v, \mathbf{P}_m)$.
It can be solved by 
introducing latent representations $\mathbf{Z}_m$
and $\mathbf{Z}_v$,
with the assumption 
that $\mathbf{Z}_v$
and $\mathbf{P}_m$ 
($\mathbf{Y}_m$
and $\mathbf{P}_v$) are conditionally independent
(the probabilistic graphical model is 
given in Figure~\ref{fig:CAEPGM}):
\begin{align}
   & p(\mathbf{Y}_m \mid \mathbf{X}_v, \mathbf{P}_v,
\mathbf{P}_m) \\
= & p(\mathbf{Z}_v \mid \mathbf{X}_v, \mathbf{P}_v, \mathbf{P}_m)
p(\mathbf{Z}_m \mid \mathbf{Z}_v, \mathbf{P}_v, \mathbf{P}_m) \nonumber\\
&p(\mathbf{Y}_m \mid \mathbf{Z}_m, \mathbf{P}_v, \mathbf{P}_m)\\
= & p(\mathbf{Z}_v \mid \mathbf{X}_v, \mathbf{P}_v)
p(\mathbf{Z}_m \mid \mathbf{Z}_v, \mathbf{P}_v, \mathbf{P}_m)
p(\mathbf{Y}_m \mid \mathbf{Z}_m, \mathbf{P}_m).
\label{eq:interp_cae}
\end{align}
Here,
the equation from (2) to (3)
is obtained from the probabilistic graphical model
of CAE shown in Figure~\ref{fig:CAEPGM},
and the removal of the condition $\mathbf{P}_m$
(from $p(\mathbf{Z}_v \mid \mathbf{X}_v, \mathbf{P}_v, \mathbf{P}_m)$
to $p(\mathbf{Z}_v \mid \mathbf{X}_v, \mathbf{P}_v)$),
and the condition $\mathbf{P}_v$
(from $p(\mathbf{Y}_m \mid \mathbf{Z}_m, \mathbf{P}_v, \mathbf{P}_m)$
to $p(\mathbf{Y}_m \mid \mathbf{Z}_m, \mathbf{P}_m)$)
from (3) to (4)
is based on the conditional independence assumption.
The three terms in (4)
correspond to three parts of our CAE:
the encoder,
the latent contextual regressor,
and the decoder, respectively.

Similarly, the latent representation alignment constraint
can be written as a conditional probability,
$ P(\mathbf{Z}_m \mid \bar{\mathbf{Z}}_m)$,
where $\bar{\mathbf{Z}}_m$ is the masked patch representations
computed from the encoder.

\vspace{1mm}
\noindent\textbf{Intuitive interpretation for the contrastive self-supervised learning.}
We consider the case in ImageNet-$1$K that the object mainly lies in the center of an image\footnote{There are a few images
in which the object does not lie in the center in ImageNet-$1$K.
The images are actually viewed as noises
and have little influence for contrastive self-supervised learning.
}.
There are $N$ randomly sampled crops from an image,
and each crop $\mathbf{I}_n$ contains a part of the center object, $\mathbf{O}_n$.
To maximize
the similarity between
two crops $\mathbf{I}_m$ and $\mathbf{I}_n$,
the pretraining 
might contain the processes:
select the regions $\mathbf{O}_m$
and $\mathbf{O}_n$
from the two crops $\mathbf{I}_m$ and $\mathbf{I}_n$,
extract their features $\mathbf{f}_{om}$
and $\mathbf{f}_{on}$,
and predict the feature of the object, $\mathbf{f}_o$,
from the part features
$\mathbf{f}_{om}$
and $\mathbf{f}_{on}$.
In this way, the features of the crops from the same image
could be similar.
Among the $N$ random crops, 
most crops contain a part of the object in the center,
and a few crops that do not contain a part of the center object
could be viewed as noises
when optimizing the contrastive loss.

After pretrained on ImageNet-$1$K
(where the object mainly lies in the center)
the encoder is able to learn the knowledge of the $1000$ classes
and localize the region containing the object belonging
to the $1000$ classes.
It is not necessary that
the object lies in the center for the testing image,
which is verified in Figure~\ref{fig:patchimportance_crop}.
This further verifies that
MoCo v3 (contrastive self-supervised pretraining)
pretrained on ImageNet-$1$K
tends to attend to the object region,
corresponding to the center region of the original image as shown in Figure~\ref{fig:patchimportance}.

\section{Experiments}
\subsection{Implementation}
We study the standard ViT small, base and large architectures,
ViT-S ($12$ transformer blocks with dimension $384$), ViT-B ($12$ transformer blocks with dimension $768$) and ViT-L ($24$ transformer blocks with dimension $1024$). 
The latent contextual regressor
consists of $4$ transformer blocks based on cross-attention
in which self-attention over masked tokens
and encoded visible patch representations
is a choice but with slightly higher computation cost and a little lower performance,
and the decoder consists of $4$ transformer blocks based on self-attention,
and an extra linear projection
for making predictions.

\subsection{Training Details}
\label{sec:trainingdetails}

\noindent\textbf{Pretraining.}
The pretraining settings are almost the same as~BEiT~\cite{bao2021beit}.
We train the CAE
on ImageNet-$1$K.
We partition the image 
of $224\times 224$
into $14\times 14$ patches
with the patch size being $16\times 16$. 
We use standard random cropping and horizontal flipping 
for data augmentation.
We use AdamW~\cite{loshchilov2017adamw} for optimization
and train the CAE for $300$/$800$/$1600$ epochs 
with the batch size being $2048$.
We set the learning rate 
as $1.5$e-$3$
with cosine learning rate decay.
The weight decay is set as $0.05$. The warmup epochs for $300$/$800$/$1600$ epochs pre-training are $10$/$20$/$40$, respectively.
We employ drop path~\cite{huang2016stochastic_depth} rate $0.1$ and dropout rate $0$.

\vspace{1mm}
\noindent\textbf{Linear probing.} 
We use the LARS~\cite{you2017large} optimizer with momentum $0.9$. 
The model is trained for $90$ epochs. The batch size is $16384$, the warmup epoch is $10$ and the learning rate is $6.4$.
Following ~\cite{he2021masked}, we adopt an extra BatchNorm layer~\cite{SergeyIoffe2015BatchNA} without affine transformation ($\texttt{affine=False}$) before the linear classifier.
We do not use mixup~\cite{HongyiZhang2017mixupBE}, cutmix~\cite{SangdooYun2019CutMixRS}, drop path~\cite{huang2016stochastic_depth}, or color jittering, and we set weight decay as zero.

\vspace{1mm}
\noindent\textbf{Attentive probing.} 
The parameters of the encoder are fixed during attentive probing.
A cross-attention module, a BatchNorm layer ($\texttt{affine=False}$), and a linear classifier are appended after the encoder.
The extra class token representation in
cross-attention
is learned as model parameters. 
The keys and the values are the patch
representations
output from the encoder. 
There is no MLP or 
skip connection operation
in the extra cross-attention module.
We use the SGD optimizer with momentum $0.9$ and train the model for $90$ epochs. The batch size is $8192$, the warmup epoch is $10$ and the learning rate is $0.4$. 
Same as linear probing,
we do not use mixup~\cite{HongyiZhang2017mixupBE}, cutmix~\cite{SangdooYun2019CutMixRS}, drop path, or color jittering, and we set weight decay as zero.

\vspace{1mm}
\noindent\textbf{Fine-tuning on ImageNet.}
We follow the fine-tuning protocol in BEiT to use layer-wise learning rate decay, weight decay and AdamW. 
The batch size is $4096$, the warmup epoch is $5$ and the weight decay is $0.05$. For ViT-S, we train $200$ epochs with learning rate $1.6$e-$2$ and layer-wise decay rate $0.75$. For ViT-B, we train $100$ epochs with learning rate $8$e-$3$ and layer-wise decay rate $0.65$.
For ViT-L, we train $50$ epochs with learning rate $2$e-$3$ and layer-wise decay rate $0.75$.

\vspace{1mm}
\noindent\textbf{Semantic segmentation on ADE$20$K.}
We use AdamW
as the optimizer. The input resolution is $512 \times 512$. The batch size is $16$.
For the ViT-B, the layer-wise decay rate is $0.65$ and the drop path rate is $0.1$. We search from four learning rates, $1$e-$4$, $2$e-$4$, $3$e-$4$ and $4$e-$4$, for all the results in Table~\ref{tab:segmentation}.
For the ViT-L, the layer-wise decay rate is $0.95$ and the drop path rate is $0.15$. We search from three learning rates for all the methods, $3$e-$5$, $4$e-$5$, and $5$e-$5$,
We conduct fine-tuning for $160$K steps. We do not use multi-scale testing.

\vspace{1mm}
\noindent\textbf{Object detection and instance segmentation on COCO.}
We utilize multi-scale training 
and resize the
image with the size
of the short side
between $480$ and $800$ 
and the long side no larger than $1333$.  
The batch size is $32$.
For the ViT-S, the learning rate is $3$e-$4$, the layer-wise decay rate is $0.75$, and the drop path rate is $0.1$. 
For the ViT-B, the learning rate is $3$e-$4$, the layer-wise decay rate is $0.75$, and the drop path rate is $0.2$. 
For the ViT-L, the learning rate is $2$e-$4$, the layer-wise decay rate is $0.8$, and the drop path rate is $0.2$.
We train the network 
with the $1 \times$ schedule:
$12$ epochs with the learning rate decayed
by $10 \times$ at epochs $9$ and $11$.
We do not use multi-scale testing.
The Mask R-CNN implementation follows MMDetection~\cite{mmdetection}.

 \begin{figure}[ht]
\centering
\includegraphics[width=.18\columnwidth]{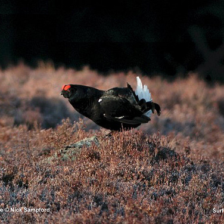}~
\includegraphics[width=.18\columnwidth]{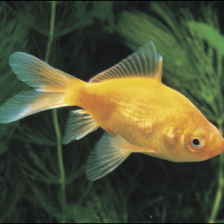}~
\includegraphics[width=.18\columnwidth]{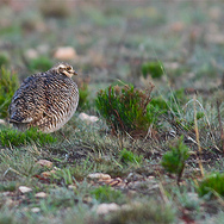}~
\includegraphics[width=.18\columnwidth]{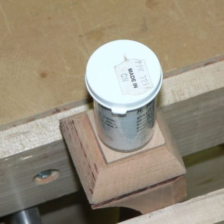}~
\includegraphics[width=.18\columnwidth]{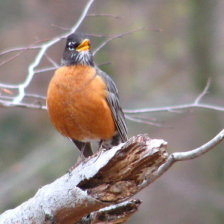}\\
\vspace{1mm}
\includegraphics[width=.18\columnwidth]{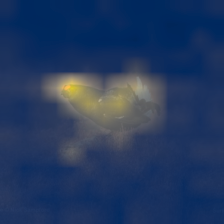}~
\includegraphics[width=.18\columnwidth]{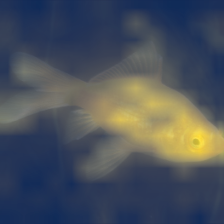}~
\includegraphics[width=.18\columnwidth]{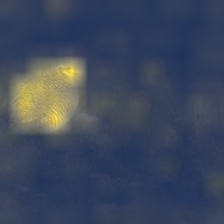}~
\includegraphics[width=.18\columnwidth]{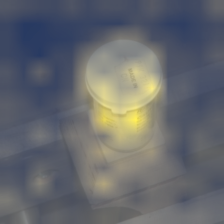}~
\includegraphics[width=.18\columnwidth]{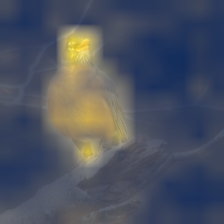}\\
\caption{
Illustrating 
the cross-attention unit
in attentive probing.
The attention map
(bottom)
is the average 
of cross-attention maps
over $12$ heads
between the extra class token and the patches.
One can see that the attended region lies mainly in the object,
which helps image classification.
}
\label{fig:attentiveprobing}
\end{figure}

\subsection{Pretraining Evaluation}
\noindent\textbf{Linear probing.}
Linear probing is widely used 
as a proxy of pretraining quality evaluation
for self-supervised representation learning.
It learns a linear classifier
over the image-level representation output from the pretrained encoder 
by using the labels of the images,
and then tests the performance
on the validation set.

\vspace{1mm}
\noindent\textbf{Attentive probing.}
The output of the encoder
pretrained with MIM methods
are representations
for all the patches.
It is not suitable
to linearly probe the representation,
averagely-pooled from patch representations,
because the image label in ImageNet-$1$K
only corresponds to a portion of patches.
It is also not suitable
to use the default class token within the encoder
because the default class token serves as
a role of aggregating the patch representations
for better patch representation extraction
and is not merely for the portion of patches
corresponding to the image label.

To use the image-level label
as a proxy of
evaluating the pretraining quality
for the encoder pretrained with MIM methods,
we need to attend the patches 
that are related to the label. 
We introduce a simple modification
by using a cross-attention unit 
with an extra class token (that is different from the class token in the encoder)
as the query
and  
the output patch representations of the encoder as the keys and the values,
followed by a linear classifier.
The introduced cross-attention unit
is able to care mainly about the patches belonging
to the $1000$ classes in ImageNet-$1$K
and remove the interference
of other patches.
Figure~\ref{fig:attentiveprobing} illustrates the effect
of the cross-attention unit,
showing that the extra cross-attention unit
can to some degree
attend the regions
that are related to the $1000$ ImageNet-$1$K classes.

\begin{table}[t]
  \centering  
  \caption{Pretraining quality evaluation
  in terms of 
  fine-tuning (FT),
  linear probing (LIN),
  and attentive probing (ATT). 
  $^\ddag$ means the number of effective epochs in~\cite{zhou2021ibot} as they adopt multi-crop augmentation (equivalently take a larger number of epochs compared to one-crop augmentation).
  We report the top-$1$ accuracy
  (in the column ATT)
  of the supervised training approach 
  DeiT~\cite{touvron2020deit}
  to show how far the ATT score is from supervised training.
 The scores
  for other models 
  and our models 
  are based on our implementations
  if not specified.
    {
    Except that * denotes using the DALL-E tokenizer, CAE adopts
    the d-VAE tokenizer trained on ImageNet-1K only.
    }
  } 
  \setlength{\tabcolsep}{6pt}
\renewcommand{\arraystretch}{1.1}
    \begin{tabular}{l c c c  c c c}
      \toprule
      Method  & \#Epochs & \#Crops & FT  & LIN & ATT \\
      \midrule
      \multicolumn{5}{l}{\emph{Methods using ViT-S}:}\\
      DeiT & $300$ &-&-&-& $79.9$  \\
      MoCo v3  & $\ \, 600^\ddag$ & $2$ & $81.7$ & $73.1$ & $ 73.8 $ \\
      BEiT   &   $300$ & $1$ & $81.7$ & $ 15.7 $ & $ 23.6 $\\
    \ours & $300$ & $1$ & $ \mathbf{82.0} $ & $ 51.8 $ & $ 65.0  $\\
      \midrule
      \multicolumn{5}{l}{\emph{Methods using ViT-B}:}\\
      DeiT & $300$ &-&-&-& $81.8$   \\
      MoCo v3   & $ \ \,  600^\ddag$ & $2$ & $83.0$ & $76.2$ & $77.0$\\
      DINO  &  $\ \,  1600^\ddag$ & $12$ & $83.3$ & $77.3$ & $ 77.8 $ %78.2
      \\
      BEiT   &  $300$ & $1$ & $ 83.0 $ & $ 37.6 $ & $ 49.4 $\\
      MAE &   $300$ & $1$ & $82.9$ & $ 61.5 $ & $71.1$ \\
      MAE &   $1600$ & $1$ & $ 83.6 $ & $ 67.8 $ & $ 74.2 $ \\
      SimMIM & $800$ & $1$ & $ 83.8 $ & $ 56.7 $ & - \\
      iBOT &  $\ \,  1600^\ddag$ & $12$ &  $83.8$ &  $79.5$ & $79.8$ \\
      \ours & $300$ & $1$ & $ 83.6 $ & $ 64.1 $ & $ 73.8 $ \\
     \ours & $800$ & $1$ & $ 83.8 $ & $ 68.6 $ & $ 75.9 $ \\
     \ours & $1600$ & $1$ & $ \mathbf{83.9} $ & $ 70.4 $ & $ 77.1 $ \\
     \oursdvae & $1600$ & $1$ & $ \mathbf{83.9} $ & $ 71.4 $ & $ 77.4 $ \\
      \midrule
      \multicolumn{5}{l}{\emph{Methods using ViT-L}:}\\
      MoCo v3$^\dagger$ & $\ \, 600^\ddag$ & $2$ & $ 84.1 $ & - & - \\
      BEiT$^\dagger$ & $1600$ & $1$ & $ 85.2 $ & - & - \\
      MAE &   $1600$ & $1$ & $ 86.0 $ & $ 76.0 $ & $ 78.8 $ \\
      \ours &   $1600$ & $1$ & $ \mathbf{86.3} $ & $ 78.1 $ & $ 81.2 $ \\
      \oursdvae &   $1600$ & $1$ & $ \mathbf{86.3} $ & $ 77.9 $ & $ 81.2 $ \\
      \bottomrule
  \end{tabular}
  \label{tab:pretrainingvaluation}
\end{table}

\vspace{1mm}
\noindent\textbf{Results.}
Table~\ref{tab:pretrainingvaluation}
shows the results 
with three schemes,
linear probing (LIN),
attentive probing (ATT),
and fine-tuning (FT)
for representative contrastive self-supervised pretraining
(MoCo v3 and DINO)
and MIM (BEiT and MAE) methods,
as well as our approach
with the targets
formed with the DALL-E tokenizer
(trained on $400$M images)
and the d-VAE tokenizer
(trained on ImageNet-$1$K without using the labels),
denoted as CAE* and CAE, respectively.
The models of MAE with $300$ epochs
and BEiT
are pretrained by us
using the official implementations,
and other models are officially released models.

We highlight a few observations.
The fine-tuning performance
for these methods are very similar
and there is only a minor difference
similar to the observation~\cite{zhou2021ibot}.
We think that
the reason is that
self-supervised pretraining 
and fine-tuning are conducted
on the same dataset
and no extra knowledge
is introduced for image classification.
The minor difference might come from the optimization aspect:
different initialization
(provided by pretrained models)
for fine-tuning.

In terms of linear probing,
the scores of the contrastive self-supervised learning methods,
MoCo v3 and DINO, are higher than the MIM methods.
This is as expected because 
contrastive self-supervised learning focuses mainly on
learning the representations for $1000$ classes
(See discussion in Section~\ref{sec:discussionAnalysis}).
The pretraining is relatively easier 
than existing MIM methods
as contrastive self-supervised learning mainly cares about the $1000$ classes
and MIM methods may care about the classes
beyond the $1000$ classes.

For the MIM methods,
the scores of attentive probing
are much larger
than linear probing.
This validates our analysis:
the MIM methods extract the representations
for all the patches,
and the classification task
needs to attend the corresponding portion of patches.

The LIN and ATT scores are similar
for contrastive self-supervised pretraining on ViT-B, e.g.,
$(76.2~\text{vs}~77.0)$
for MoCo v3
and $(77.3~\text{vs}~77.8)$ for DINO.
This means that the extra cross-attention in attentive probing
does not make a big difference,
which is one more evidence for our analysis in Section~\ref{sec:discussionAnalysis} that
they already focus
mainly 
on the region where the instance in the $1000$ categories lies.

\subsection{Downstream Tasks}

\begin{table}[t]
    \centering
    \caption{Semantic segmentation on ADE$20$K.
    All the results are based on the same
    implementation
    for semantic segmentation. \#Epochs refers to the number of pretraining epochs.
    $^\ddag$ means the number of effective epochs in~\cite{zhou2021ibot} as the method uses multi-crop pretraining augmentation (See Table~\ref{tab:pretrainingvaluation}).
    SplitMask~\cite{el2021large} is pretrained on ADE20K for 21000 epochs.
    { 
    $^\dagger$: these results are from \cite{he2021masked}.
    }
    } 
    \setlength{\tabcolsep}{23pt}
\renewcommand{\arraystretch}{1.1}
        \begin{tabular}{l c c}
        \toprule
        Method  
        & \#Epochs  & mIoU  \\
         \midrule
      \multicolumn{3}{l}{\emph{Methods using ViT-B}:}\\
        SplitMask & -- & $45.7$ \\
        BEiT & $300$ &  $45.5$ \\
        BEiT & $800$ &  $ 46.5 $   \\ 
        mc-BEiT & 800 & $47.0$ \\
        DeiT & $300$ &  $47.0$ \\
        MoCo v3 & $\ \, 600^\ddag$ &  $ 47.2 $ \\
        DINO & $\ \, 1600^\ddag$ &  $ 47.2 $ \\
        MAE & $300$ & $45.8$\\
        MAE & $1600$ &  $48.1$\\
        Ge$^2$-AE & $800$ & $48.9$ \\
        A$^2$MIM & $800$ & $49.0$ \\
         iBOT & $\ \, 1600^\ddag$ & $50.0$ \\
        \ours & $300$ &  $ {48.3} $ \\ 
        \ours & $800$  &  $ 49.7 $ \\
        \ours & $1600$ &  $ \mathbf{50.2} $ \\
        \oursdvae & $1600$ &  $ 50.1 $ \\
    \midrule
      \multicolumn{3}{l}{\emph{Methods using ViT-L}:}\\
      MoCo v3$^{\dagger}$ & $\ \, 600^\ddag$  & $49.1$\\
      BEiT$^\dagger$ & $1600$  & $53.3$\\
      MAE & $1600$  & $53.6$\\
      \ours & $1600$  &  $ \mathbf{54.7} $ \\
      \oursdvae & $1600$  &  $ 54.6 $ \\
        \bottomrule     
    \end{tabular} 
    \label{tab:segmentation}
\end{table}

\begin{table*}[t]
    \centering
    \caption{Object detection and instance segmentation on COCO. 
    Mask R-CNN is adopted
    and trained with the $1\times$ schedule.
    All the results are based on
    the same implementation
    for object detection and instance segmentation.
    \#Epochs refers to the number of pretraining epochs on ImageNet-$1$K.
    $^\ddag$ means the number of effective epochs in~\cite{zhou2021ibot} (See Table~\ref{tab:pretrainingvaluation}).
    }
\setlength{\tabcolsep}{9.8pt}
\renewcommand{\arraystretch}{1}
    \small
\begin{tabular}{l c c c  c c c   c  c c}
        \toprule
        \multirow{2}{*}{Method} & 
        \multirow{2}{*}{\#Epochs} & 
        \multirow{2}{*}{Supervised} & 
        \multirow{2}{*}{Self-supervised} &
        \multicolumn{3}{c}{Object detection}& 
        \multicolumn{3}{c}{Instance segmentation}\\ 
        \cline{5-10}
          &  &  &   &{ $\text{AP}^{b}$} & {$\text{AP}^{b}_{50}$} & {$\text{AP}^{b}_{75}$} & 
        {$\text{AP}^{m}$} & 
        {$\text{AP}^{m}_{50}$} & 
        {$\text{AP}^{m}_{75}$}\\ 
        \midrule
      \multicolumn{9}{l}{\emph{Methods using ViT-S}:}\\
        DeiT  & $300$ &\cmark & \xmarkg  & $ 43.1 $ & $ 65.2 $ & $ 46.6 $ & $ 38.4 $ & $ 61.8 $ & $ 40.6 $ \\
        MoCo v3  & $\ \, 600^\ddag$ & \xmarkg & \cmark &  $ 43.3 $ & $ 64.9 $ & $ 46.8 $ & $ 38.8 $ & $ 61.6 $ & $ 41.1 $ \\
        BEiT & $300$ &\xmarkg & \cmark & $ 35.6 $ & $ 56.7 $ & $ 38.3 $ & $ 32.6 $ & $ 53.3 $ & $ 34.2 $ \\
        \ours  & $300$ &    \xmarkg & \cmark &  $ \mathbf{44.1} $ & $ 64.6 $ & $ 48.2 $ & $ \mathbf{39.2} $ & $ 61.4 $ & $ 42.2 $ \\
        \midrule
      \multicolumn{9}{l}{\emph{Methods using ViT-B}:}\\
        DeiT  & $300$ &\cmark & \xmarkg  & $ 46.9 $ & $ 68.9 $ & $ 51.0 $ & $ 41.5 $ & $ 65.5 $ & $ 44.4 $ \\
        MoCo v3 & $\ \, 600^\ddag$ & \xmarkg & \cmark & $ 45.5 $ & $67.1$ & $ 49.4 $ & $ 40.5 $ & $ 63.7 $ & $ 43.4 $ \\
        DINO & $\ \, 1600^\ddag$  & \xmarkg & \cmark &  $ 46.8 $ & $ 68.6 $ & $ 50.9 $ & $ 41.5 $ & $ 65.3 $ & $ 44.5 $ \\
        BEiT & $300$  &\xmarkg & \cmark & $ 39.5 $ & $ 60.6 $ & $ 43.0 $ & $ 35.9 $ & $ 57.7 $ & $ 38.5 $ \\
        BEiT & $800$   &\xmarkg & \cmark & $ 42.1 $ & $ 63.3 $ & $ 46.0 $ & $ 37.8 $ & $ 60.1 $ & $ 40.6 $ \\
        MAE & $300$    & \xmarkg & \cmark &  $ 45.4 $ & $ 66.4 $ & $ 49.6 $ & $ 40.6 $ & $ 63.4 $ & $ 43.7 $ \\
        MAE & $1600$    & \xmarkg & \cmark &  $ 48.4 $ & $ 69.4 $ & $ 53.1 $ & $ 42.6 $ & $ 66.1 $ & $ 45.9 $ \\
        iBOT & $\ \, 1600^\ddag$    & \xmarkg & \cmark &  $ 48.2 $ & $ 69.7 $ & $ 52.8 $ & $ 42.7 $ & $ 66.5 $ & $ 46.0 $ \\
        \ours & $300$  & \xmarkg & \cmark &  $ 48.4 $ & $ 69.2 $ & $ 52.9 $ & $ 42.6 $ & $ 66.1 $ & $ 45.8 $ \\
        \ours & $800$  & \xmarkg & \cmark &  $ 49.8 $ & $ 70.7 $ & $ 54.6 $ & $ 43.9 $ & $ 67.8 $ & $ 47.4 $ \\
        \ours & $1600$  & \xmarkg & \cmark &  $ 50.0 $ & $ 70.9 $ & $ 54.8 $ & $ 44.0 $ & $ 67.9 $ & $ 47.6 $ \\
        \oursdvae & $1600$  & \xmarkg & \cmark &  $ \mathbf{50.2} $ & $ 71.0 $ & $ 54.9 $ & $ \mathbf{44.2} $ & $ 68.3 $ & $ 47.9 $ \\
        \midrule
      \multicolumn{9}{l}{\emph{Methods using ViT-L}:}\\
        MAE & $1600$  & \xmarkg & \cmark &  $ 54.0 $ & $ 74.3 $ & $ 59.5 $ & $ 47.1 $ & $ 71.5 $ & $ 51.0 $ \\
        \ours & $1600$  & \xmarkg & \cmark &  $ 54.5 $ & $ 75.2 $ & $ 60.1 $ & $ 47.6 $ & $ 72.2 $ & $ 51.9 $ \\
        \oursdvae & $1600$  & \xmarkg & \cmark &  $ \mathbf{54.6} $ & $ 75.2 $ & $ 59.9 $ & $ \mathbf{47.6} $ & $ 72.0 $ & $ 51.9 $ \\
        \bottomrule     
    \end{tabular} 
    \vspace{-0.2cm}
    \label{tab:cocodetection}
\end{table*}

\noindent\textbf{Semantic segmentation on ADE$\mathbf{20}$K}~\cite{zhou2017scene}\textbf{.}
We follow the implementation~\cite{bao2021beit}
to use UperNet~\cite{xiao2018unified}.
The CAE with the tokenizers learned over ImageNet-$1$K
performs almost the same as the tokenizers learned over $400$M images provided by DALL-E (CAE*),
implying that the tokenizer trained
on ImageNet-$1$K (without using the labels) or a larger dataset
does not affect the pretraining quality and
accordingly the downstream task performance.

Table~\ref{tab:segmentation} shows that using the ViT-B, 
our CAE* with $300$ training epochs
performs better
than DeiT,
MoCo v3, DINO,
MAE ($1600$ epochs)
and BEiT. 
Our CAE* ($1600$ epochs)
further improves the segmentation scores
and outperforms MAE ($1600$ epochs), MoCo v3 and
DeiT by $2.1$, $3.0$ and $3.2$, respectively. Using ViT-L, our CAE* ($1600$ epochs) outperforms BEiT ($1600$ epochs) and MAE ($1600$ epochs) by $1.4$ and $1.1$, respectively.

The superior results over supervised and contrastive self-supervised pretraining methods,
DeiT,
MoCo v3 and DINO,
stem from 
that our approach captures the knowledge beyond the $1000$ classes in ImageNet-$1$K.
The superior results over BEiT
and MAE stems from that
our CAE makes predictions 
in the encoded representation space
and that representation learning
and pretext task completion
are separated.

\begin{table*}[t]
  \centering
  \caption{Top-1 classification accuracy on the Food-$101$, Clipart and Sketch datasets. The backbone is ViT-B.}
  \setlength{\tabcolsep}{21pt}
\renewcommand{\arraystretch}{1.1}
        \begin{tabular}{l  c  c  c  c c }
            \toprule
            {Method} & {Supervised} & {Self-supervised} & Food-$101$ & Clipart  & Sketch \\
              \hline
              Random Init. &   \xmarkg &  \xmarkg & 82.77 & $52.90$ &  $46.42$ \\
              DeiT & \cmark &  \xmarkg & $91.81$ & $81.18$ &  $73.45$ \\
              DINO &  \xmarkg & \cmark & $91.67$ & $80.72$  &   $73.13$  \\
              MAE & \xmarkg & \cmark & $93.19$ & $80.63$  &    $73.87$ \\
             \ours & \xmarkg & \cmark & $\mathbf{93.32}$ & $\mathbf{81.84}$  & $\mathbf{74.65}$ \\
            \bottomrule
        \end{tabular} 
        \label{tab:more_classification}
\end{table*}

\begin{table*}[t]
\caption{Ablation studies
for the decoder and the alignment constraint 
in our CAE. 
All the models are pretrained on ImageNet-$1$K with $300$ epochs.
}
\label{tab:ablation}
\setlength{\tabcolsep}{10pt}
\renewcommand{\arraystretch}{1.1}
\centering
\begin{tabular}{ccccccccc}
\toprule
 Decoder & Alignment & LIN & ATT & FT  &  ADE Seg.  &  COCO Det. & \#Params & Training Time \\
\midrule
 \xmarkg & \xmarkg & $60.3$ & $71.2$ & $82.9$ & $ 47.0 $ &  $46.9$ & $120.32$ M & $1 \times$ \\
 \cmark & \xmarkg & $63.1$ & $72.7$ & $83.4$ & $ 47.1 $ &  $47.2$ & $148.68$ M & $1.14 \times$ \\
 \xmarkg & \cmark & $62.0$ & $71.5$ & $83.4$ & $47.1$ &  $47.2$ & $120.32$ M & $1.12 \times$ \\
 \cmark & \cmark & $64.1$ & $ 73.8 $ & $83.6$ & $ 48.3 $ &  $ 48.4 $ & $148.68$ M & $1.24 \times$\\
\bottomrule
\end{tabular} 
\end{table*}

\begin{table}[t]
    \centering
    \caption{
    The results of object detection and instance segmentation on COCO with the Cascaded Mask-RCNN framework ($1\times$ schedule). 
    ViT-B is used for all experiments.
    All the detection results
    are from our implementation.
    } 
    \setlength{\tabcolsep}{13pt}
\renewcommand{\arraystretch}{1.1}
    \small
        \begin{tabular}{l c c c}
        \toprule
            \multirow{1}{*}{Method} & \multirow{1}{*}{\#Epochs} &
            $\text{AP}^{b}$ & $\text{AP}^{m}$ \\
        \midrule
        MAE~\cite{he2021masked}  & $1600$ & $51.3$ & $44.3$ \\
        mc-BEiT~\cite{zhou2021ibot}   & $800$ &  $50.1$ & $43.1$ \\
        iBOT~\cite{zhou2021ibot}  & $1600$  &  $51.2$ & $44.2$ \\ 
        \ours  & $300$  & $51.6$ & $44.6$ \\
        \ours  & $800$  & $52.8$ & $45.5$  \\
        \ours  & $1600$ & $\mathbf{52.9}$ & $\mathbf{45.5}$  \\
        \bottomrule     
    \end{tabular} 
    \label{tab:det_cascade}
\end{table}

 \vspace{1mm}
\noindent\textbf{Object detection and instance segmentation on COCO}~\cite{lin2014microsoft}\textbf{.}
We adopt the Mask R-CNN approach~\cite{he2017mask}
that produces bounding boxes and instance masks simultaneously,
with the ViT as the backbone.
The results are given in Table~\ref{tab:cocodetection}.
We report the box AP for object detection and the mask AP for
instance segmentation.
The observations are consistent with those for semantic segmentation
in Table~\ref{tab:segmentation}.
Our CAE* ($300$ epochs, ViT-B)
is superior to
all the other models except 
that a little lower than
MAE ($1600$ epochs).
Our approach ($1600$ epochs)
outperforms MAE ($1600$ epochs),
MoCo v3 and
DeiT by $1.6$, $4.5$ and $3.1$, respectively. Using ViT-L, our CAE achieves $54.6$ box AP and outperforms MAE by $0.6$.

We also report the results of object detection and instance segmentation on COCO with the Cascaded Mask R-CNN framework~\cite{ZhaoweiCai2021CascadeRH} in Table~\ref{tab:det_cascade}. Results show that our CAE performs better than other methods.

In addition, we conduct experiments on the scaling ability of CAE on the detection task. The detection model is built upon ViT-Huge~\cite{DosovitskiyB0WZ21}, DINO~\cite{HaoZhang2023DINODW}, and Group DETR~\cite{QiangChen2022GroupDF}
(see~\cite{groupdetrv2} for more details). The ViT-Huge is pretrained on ImageNet-$22$K~\cite{deng2009imagenet} using CAE. 
We are the first to obtain $64.6$ mAP on COCO \textit{test-dev}, 
which outperforms previous methods with larger models and more training data (e.g., BEIT-3~\cite{WenhuiWang2023ImageAA} ($63.7$ mAP) and SwinV2-G~\cite{ZeLiu2021SwinTV} ($63.1$ mAP)).

\begin{table}[t]
  \centering
  \caption{The effect of mask ratios. The backbone is ViT-B. Models are trained for $300$ epochs.} 
  \setlength{\tabcolsep}{15pt}
\renewcommand{\arraystretch}{1.1}
        \begin{tabular}{ c  c  c c }
            \toprule
            {Mask Ratio} & {LIN} & {ATT} & ADE Seg \\
              \hline
              $40\%$ & $63.1$ & 	$73.0$	& $47.2$ \\
              $50\%$ & $64.1$ & 	$73.8$	& $48.3$ \\
              $60\%$ & $64.8$ & 	$74.2$	& $48.1$ \\
            \bottomrule
        \end{tabular} 
        \label{tab:mask_ratio}
\end{table}

\vspace{1mm}
\noindent\textbf{Classification.} 
We conduct fine-tuning experiments on three datasets: Food-$101$~\cite{bossard14}, Clipart~\cite{castrejon2016learning}, and Sketch~\cite{castrejon2016learning}. Results in Table~\ref{tab:more_classification} show that the proposed method outperforms the previous supervised method (DeiT) and self-supervised methods (DINO, MAE).

\subsection{Ablation Studies}

\noindent \textbf{Decoder and alignment.}
The CAE architecture contains
several components for pretraining the encoder:
regressor
and alignment for masked representation prediction,
decoder with a linear layer for masked
patch reconstruction.
We observe that
if the pretraining task,
masked patch reconstruction, is not included,
the training collapses, leading to a trivial solution. 
We thus study
the effect of 
the decoder
(when the decoder is removed, we
use a linear layer to predict the targets),
which is helpful for target reconstruction,
and the alignment,
which is helpful for representation prediction.

Table~\ref{tab:ablation} shows the ablation results.
We report the scores for 
linear probing,
attentive probing, fine-tuning
and downstream tasks: semantic segmentation on the ADE$20$K dataset
and object detection on COCO
with the DALL-E tokenizer as the target.
One can see that 
the downstream task performance is almost the same
when only the decoder is added
and that the performance
increases
when the decoder and the alignment are both added.
This also verifies that
the alignment is important
for ensuring that
the predicted representations 
of masked patches
lie in the encoded representation space
and thus the predictions are made in the encoded representation space,
and accordingly improving the representation quality. 
Without the decoder, 
the performance drops. 
This is because the reconstruction from the semantic representation
to the low-level targets
cannot be done through a single linear layer,
and no decoder will deteriorate the semantic quality of the encoder.
The additional computational cost, i.e. the number of parameters and training time, brought by the decoder and alignment is relatively small, e.g., increasing the number of parameters to $1.23\times$ and training time to $1.24\times$.

\vspace{1mm}
\noindent \textbf{Mask ratio.}
We also conduct experiments with different mask ratios including $40\%$, $50\%$, and $60\%$. Results are listed in Table~\ref{tab:mask_ratio}. We find that ratio $50\%$ gets better results than ratio $40\%$. Adopting a higher mask ratio ($60\%$) could further improve the performance of linear probing and attentive probing, while the semantic segmentation performance is reduced by $0.2$\%. We choose $50\%$ in our work unless specified.

\begin{table}[t]
  \centering
  \caption{The effect of reconstruction targets on the performance of CAE. The backbone is ViT-B. Models are trained for $1600$ epochs.} 
  \setlength{\tabcolsep}{11pt}
\renewcommand{\arraystretch}{1.1}
        \begin{tabular}{ c  c  c c }
            \toprule
            {Targets} & {LIN} & {ATT} & ADE Seg \\
              \hline
              DALL-E tokenizer & $70.4$ & 	$77.1$	& $50.2$ \\
              d-VAE tokenizer & $71.4$ & 	$77.4$	& $50.1$ \\
              RGB pixel value & $72.4$ & 	$77.0$	& $50.4$ \\
            \bottomrule
        \end{tabular} 
        \label{tab:abla_target}
\end{table}

\vspace{1mm}
\noindent\textbf{\#layers in the regressor and decoder.} 
For the number of layers in the latent contextual regressor and decoder, we tried four choices: $1$-layer, $2$-layers, $4$-layer, and $5$-layer. The results for linear probing are $58.7$, $62.1$, $64.1$, and $64.2$. The results for attentive probing are $67.5$, $71.1$, $73.8$, and $73.7$.
We empirically observed that $4$-layer outperforms other choices overall.

\vspace{1mm}
\noindent\textbf{Loss tradeoff parameter.}
There is a tradeoff variable $\lambda$
in the loss function given in Equation~\ref{eqn:lossfunction}.
We did not do an extensive study
and only tried three choices, 
$\lambda =1$, $\lambda =1.5$ and $\lambda =2$.
The linear probing results are $63.4$, $63.7$ and $64.1$, respectively.
The choice $\lambda =1$ works also well, slightly worse than $\lambda =2$ that is adopted in our experiment.

\vspace{1mm}
\noindent\textbf{Reconstruction targets.}
To study the impact of different pretraining targets on model performance, we conduct additional experiments on the RGB pixel value target. Comparing the results with DALL-E tokenizer and d-VAE tokenizer trained on ImageNet-1K, the model shows better linear probe and segmentation results but inferior in attentive probe, as shown in Table~\ref{tab:abla_target}. Pretraining with these three targets obtains similar performance, illustrating that CAE does not rely on specific pretraining targets.

\section{Conclusion}
The core design
of our CAE architecture 
for masked image modeling
is that predictions are made
from visible patches
to masked patches
in the encoded representation space.
We adopt two pretraining tasks:
masked representation prediction
and masked patch reconstruction.
Experiments demonstrate
the effectiveness of the CAE design.
In addition,
we also point out that
the advantage of MIM methods
over typical contrastive self-supervised pretraining
and supervised pretraining on ImageNet-$1$K
is that MIM learns the representations
for all the patches,
while 
typical contrastive self-supervised pretraining (e.g., MoCo and SimCLR)
and supervised pretraining
tend to learn semantics
mainly from center patches
of the original images
and little from non-center patches.

Possible extensions, 
as mentioned in the arXiv version~\cite{CAE2022},
include:
investigating the possibility only 
considering the pretraining task,
masked representation prediction,
without masked patch reconstruction,
pretraining a depth-wise convolution network with masked convolution,
and pretraining with the CLIP targets~\cite{CAEv22022}.

\noindent \textbf{Potential limitations.} 
The proposed method may face challenges when dealing with
large and contiguous masked regions in an image,
e.g., the whole object region is almost masked. Obtaining plausible and high-quality reconstruction for large areas can be particularly difficult, as the model has to infer the missing information based on limited available context. This is a common limitation of Masked Image Modeling methods, and our proposed method is not exempt from it.

\section*{Acknowledgments}
We would like to acknowledge Hangbo Bao, Xinlei Chen, Li Dong, Qi Han, Zhuowen Tu, Saining Xie, and Furu Wei for the helpful discussions.

\section*{Declarations}
\begin{itemize}
\item Funding

This work is partially supported by the National Key Research and Development
Program of China (2020YFB1708002), National Natural Science Foundation of China (61632003, 61375022, 61403005), Grant SCITLAB-20017 of Intelligent Terminal Key Laboratory of SiChuan Province, Beijing Advanced Innovation Center for Intelligent Robots and Systems (2018IRS11), and PEK-SenseTime Joint Laboratory of Machine Vision. Ping Luo is supported by the General Research Fund of HK No.27208720, No.17212120, and No.17200622.

\item Code availability 

Our code will be available at \url{https://github.com/Atten4Vis/CAE}.

\item Availability of data and materials

The datasets used in this paper are publicly available. 
ImageNet: \url{https://www.image-net.org/},\\ ADE$20$K: \url{https://groups.csail.mit.edu/vision/datasets/ADE20K/}, \\
COCO: \url{https://cocodataset.org/}, \\
{Food-101}: \url{https://data.vision.ee.ethz.ch/cvl/datasets_extra/food-101/}, \\
{Clipart}: \url{http://projects.csail.mit.edu/cmplaces/download.html},\\
{Sketch}: \url{http://projects.csail.mit.edu/cmplaces/download.html}.

\end{itemize}

\bibliographystyle{plain}
\bibliography{_sn-bibliography}

\end{document}